\documentclass[letterpaper,journal]{IEEEtran}
\usepackage{amsmath,amsfonts}
\usepackage{array}
\usepackage[caption=false,font=footnotesize]{subfig}
\usepackage{textcomp}
\usepackage{stfloats}
\usepackage{url}
\usepackage{verbatim}
\usepackage{graphicx}
\usepackage{cite}
\usepackage{amssymb, amsthm}
\usepackage{physics}
\usepackage{mathtools}
\DeclarePairedDelimiterX{\infdivx}[2]{(}{)}{%
  #1\;\delimsize\|\;#2%
}

\NewDocumentEnvironment{alignb}{b}{%
  \begin{align*}
  \refstepcounter{equation} #1 \tag{\theequation}
  \end{align*}
}{} 
\usepackage{bm} 
\usepackage{dsfont} 
\newcommand{\ind}{\mathds{1}}  
\DeclareMathAlphabet{\mymathbb}{U}{BOONDOX-ds}{m}{n}
\usepackage{cite}
\usepackage[noend,lined,boxed,ruled,longend]{algorithm2e} 
\usepackage{multicol}
\usepackage{color}
\usepackage{multirow} 
\usepackage{graphicx}
\graphicspath{{./Fig/}}
\usepackage[dvipsnames]{xcolor} 
\usepackage{hyperref}
\hypersetup{
    colorlinks=false,
    linkcolor=NavyBlue,
    filecolor=OrangeRed,      
    urlcolor=Emerald,
    citecolor=Emerald,
    pdftitle={Pipecycle-EHFL},
    }
\usepackage{apptools} 
\usepackage{centernot} 
\usepackage{acronym}
\AtAppendix{\counterwithin{lemma}{section}}

\newtheorem{assumption}{Assumption}
\newtheorem{corollary}{Corollary}
\newtheorem{lemma}{Lemma}

\newtheorem{theorem}{Theorem}

\SetFuncSty{textsc}
\SetKwProg{Fn}{Function}{:}{end}
\SetKwInput{KwData}{Input}
\SetKwInput{KwResult}{Output}
\SetKwProg{Init}{Initialize}{}{}
\SetKwFor{ParFor}{for}{do in parallel}{end for}
\SetKwRepeat{Do}{do}{while}
\SetCommentSty{itshape} 
\SetCommentSty{mycommfont}

\DeclareMathOperator{\Bin}{Binomial} 

\begin{document}

\title{Computation-aware Energy-harvesting Federated Learning with Pipelined Cyclic Scheduling}

\author{Eunjeong Jeong,~\IEEEmembership{Member,~IEEE,}
        Nikolaos Pappas,~\IEEEmembership{Senior Member,~IEEE}
\thanks{E. Jeong and N. Pappas are with the Department of Computer and Information Science, Linköping University, Sweden. e-mail: \{eunjeong.jeong, nikolaos.pappas\}@liu.se.

The manuscript expands the contribution in the authors' previous work~\cite{jeong25:FedBacys} presented in IEEE 26th International Workshop on Signal Processing and Artificial Intelligence for Wireless Communications (SPAWC).

This work was supported by ELLIIT,  ROBUST-6G (grant no. 101139068), and 6G-LEADER (grant no. 101192080).
The computations and data handling were enabled by resources provided by the National Academic Infrastructure for Supercomputing in Sweden (NAISS), partially funded by the Swedish Research Council through grant agreement no. 2022-06725.
}}



\maketitle

\begin{abstract}
Federated learning (FL) is a powerful paradigm for distributed learning, but increasing model complexity leads to significant energy consumption from client-side computations for local training. This challenge is critical in energy-harvesting FL (EHFL) systems, where the participation availability of each device fluctuates because of limited energy. To address this, we propose PipeCycle, a battery-aware distributed learning framework that organizes clients into pipelined cyclic groups. When a group completes its intra-group aggregation, its aggregated model is relayed directly to a newly formed group as a reference for local training, allowing multiple groups to coexist in the pipeline while overlapping client recharging periods with active training in other pipeline stages. We provide a convergence analysis of PipeCycle under a realistic energy consumption model in which local training spans multiple time slots, and show that the cyclic structure of the pipeline imposes a finite-horizon staleness bound that avoids the exponential factors typical of asynchronous FL analyses. Numerical experiments across both IID and non-IID data and various battery charging probabilities show that \textit{PipeCycle reaches a target accuracy with substantially lower cumulative energy than existing FL baselines, particularly under severe label skew where competing cyclic schemes collapse to near-chance accuracy}.
\end{abstract}


\section{Introduction}\label{sect:intro}

\acrodef{fl}[FL]{federated learning}
\acrodef{eh}[EH]{energy-harvesting}
\acrodef{ehfl}[EHFL]{energy-harvesting federated learning}
\acrodef{sgd}[SGD]{stochastic gradient descent}

\Ac{fl} \cite{konevcny16:FL} is a distributed optimization framework that has seen rapid growth thanks to enabling privacy-preserving collaborative learning. Ensuring sustainable \ac{fl} has become critical as intelligent services are increasingly deployed on battery-powered edge devices~\cite{prabakeran2025}. As training models grow in complexity, client-side computations have taken the dominant portion of energy usage across all stages \cite{yousefpour23:greenFL, thakur25:greenFL}, highlighting the need to optimize user scheduling to balance \ac{fl} performance against energy efficiency~\cite{salh23, wiesner24:fedzero}.

Despite these concerns, few works fully integrate a comprehensive energy consumption model into their optimization frameworks. While some focus on communication-efficient techniques such as data compression~\cite{li25:fedbif} or computation-efficient methods like adaptive local training~\cite{wang24:fast, bian25:fedalt}, they often neglect the constraints imposed by battery-aware dynamics in real-world systems. Each local computation and uplink transmission depletes energy from a finite battery, which requires time to recharge \cite{cheikh25, ye24}. This gap underscores the demand for algorithms that explicitly integrate energy availability and battery dynamics into the optimization process.

In response, \ac{ehfl} \cite{liu21:iccc, chen22, shen22, hamdi22:ehfl, zeng24:ehfl, an24:ehfl, bagci25} has emerged as a promising paradigm, enabling clients to replenish energy from external sources. However, the current \ac{ehfl} literature commonly assumes that local training completes within a single, short time slot between two adjacent global communication rounds \cite{wu25, gu21:mifa, cho23}. This assumption is increasingly unrealistic as models grow in size and complexity. A more practical formulation is therefore needed to account for the substantial time and energy required by local updates.

In this work, we address these challenges by proposing PipeCycle, a battery-aware pipelined cyclic client participation framework for \ac{ehfl} networks. PipeCycle clusters users based on their remaining battery levels and allows multiple groups to coexist in pipelines, training at different stages concurrently. In this mechanism, groups relay their aggregated models: when a group completes intra-group aggregation, its hub directly passes the aggregated model to a newly formed group as the initial reference for its local training
while the server adopts the same model in parallel. This relay chain overlaps the recharging periods of inactive clients with active training in other pipeline stages, and the cyclic structure imposes a bounded staleness on the reference models propagating through the pipeline.

Beyond the algorithmic design, we identify a fundamental trade-off between the convergence speed and the inference capability, both of which are governed by the maximum number of concurrent groups $G$. Allowing more groups to coexist lets the system complete more aggregation events over the same training horizon, which speeds up learning early on. However, when the user population is fixed, larger $G$ means fewer clients in each group; each aggregated update experiences more client drifts, and the model settles at a less accurate point in the long run. Our convergence analysis captures both sides of this trade-off and yields a closed-form expression for an optimal number of concurrent groups $G^\star$ in terms of the local training cost and other system parameters.

Through numerical experiments, we validate PipeCycle across both IID and non-IID data and a range of battery charging probabilities. The experiments show that the value of $G$ that performs best depends on the charging probability and the data distribution, in the direction predicted by our analysis. Furthermore, PipeCycle reaches a target accuracy with substantially lower cumulative energy than existing \ac{fl} baselines, particularly under severe label skew where competing cyclic schemes collapse to near-chance accuracy.

\section{Background and Related Work}\label{sect:bkgrnd}

\subsection{Energy-Harvesting and Intermittent Availability in FL}
The study of \ac{fl} in resource-constrained environments has evolved to address the intermittent availability of clients, which is a common challenge in large-scale wireless networks~\cite{guler21}. This intermittent participation can be caused by various factors, including randomness in data arrivals~\cite{hu25-tcom}, straggler problems~\cite{zheng25-tcom}, unreliable channels~\cite{li23:redundancy-tmc, ko23-tmc}, heterogeneous computation capabilities~\cite{dong24-tmc, pang23:incentive-tmc}, or energy deficiency~\cite{chen23-tcom} as in our scenario. Research on intermittent \ac{fl}~\cite{gu21:mifa, ribero23} provides valuable insights into client sampling and adaptive strategies for handling sporadic participation.

Building on this, the more specialized field of \ac{ehfl} has emerged, where client availability is directly linked to their ability to collect and manage energy. Existing \ac{ehfl} studies follow three main directions. The first maximizes the number of participants per round to reduce training loss~\cite{liu21:iccc, shen22, bagci25, wu25}, treating richer participation as the main lever on convergence. The second minimizes transmission overhead by compressing or adapting the per-round uplink payload to fit a tight energy budget~\cite{valentedasilva25:FLDA, li23:anycostfl, bouzinis23}. The third jointly optimizes client scheduling and radio-resource allocation against the underlying energy-causality constraint, rather than acting on participation or payload in isolation~\cite{dasilva21, chen22, zhang26:twc}. A common assumption across all three is that energy arrivals follow a predetermined probabilistic distribution and, more critically, that a client completes its \emph{local training within a single, brief time slot} between two adjacent communication events. This neglects the significant time and energy costs of local computation, which is a limitation our work addresses.

\subsection{Cyclic Client Participation}\label{subsect:cyclic-parti}
Cyclic client participation has been explored as a solution to enhance \ac{fl} performance. Grouping and having clients participate sequentially achieve faster asymptotic convergence rates compared to vanilla FedAvg~\cite{cho23, crawshaw24}. Selecting clients based on participation frequency can accelerate convergence at a rate of $\mathcal{O}(1/VT)$, where $V$ is a constant determined by data heterogeneity and partial client participation, and $T$ is the number of communication rounds~\cite{zhu23}.

Beyond convergence speed, cyclic participation can alleviate the impact of data heterogeneity by clustering users based on distribution standards~\cite{sattler21, kim21, li22, ghosh22:ifca}. This clustering-based line of work remains active: recent studies pair cluster formation with convergence analysis and resource optimization~\cite{xu24:iotj}, with participation-aware client selection~\cite{huang24}, and with robustness to distributional shift across clients~\cite{chen24:fedccfa}.

In this work, we emphasize that grouping users in cyclic participation provides two complementary benefits for energy-harvesting devices: (1) aggregating updates within groups can stabilize the number of active participants per communication round, and (2) partitioning the user pool across groups allows subsequent groups to recharge while others train. Our proposed scheme leverages these benefits to reduce performance fluctuations through a controlled number of participants per round, while cutting energy consumption by avoiding redundant computation in a battery-constrained environment.
\section{System Model}\label{sect:system_model}

We consider the following optimization task over $N$ clients whose goal is to minimize
\begin{align}
    f(\mathbf{x}) := \frac{1}{N}\sum_{i\in[N]} f_i(\mathbf{x}),
    \label{fn:obj}
\end{align}
where $\mathbf{x}\in \mathbb{R}^{d_\text{FL}}$ is a $d_\text{FL}$-dimensional model parameter, $f_i$ is a local loss function, and $[N]=\{1,\cdots,N\}$ is the set of network users. Each client possesses a local training dataset $\mathcal{D}^{(i)}$ that is not shared with other clients.

\Ac{ehfl} follows the standard \ac{fl} workflow composed of server-side computation, downlink communication, client-side computation, and uplink communication. Separated from the communication and computation stages, each client in \ac{ehfl} independently harvests and stores energy in an energy queue, which is modeled similarly to \cite{shen22}. We assume normalized energy units, such that an activity taking one time slot requires one battery unit to be spent. The system operates in discrete time slots, with each unit of time corresponding to a single scheduling interval.

\begin{figure}[t]
    \centering
    \includegraphics[width=0.85\linewidth]{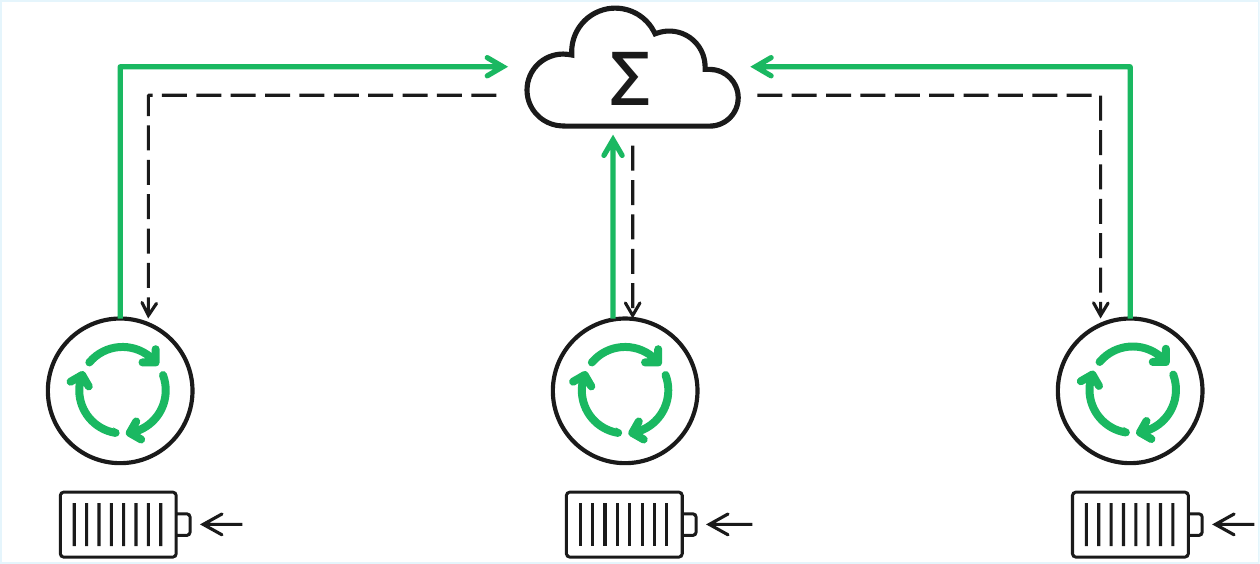}
    \caption{A schematic view of \ac{ehfl}. Steps in green require sufficient battery for operation.}
    \label{fig:ehfl-schematic}
\end{figure}

A schematic view of \ac{ehfl} is depicted in Fig.~\ref{fig:ehfl-schematic}. In each time slot, a user that is not currently engaged in a previous task can decide to perform one of three actions: (1) transmitting a message, (2) training its local model, or (3) staying idle. Transmission and idle actions each occupy a single slot, while a client that initiates local model training remains busy for the subsequent $B$ time slots and cannot make further decisions until this task is complete.

Let $\mathcal{T}_s^{(i)}\in\{0,1\}$ and $\mathcal{L}_s^{(i)}\in\{0,1\}$ be binary indicators such that $\mathcal{T}_s^{(i)}=1$ if user $i$ transmits at slot $s$, and $\mathcal{L}_s^{(i)}=1$ if user $i$ is performing local model training at slot $s$. Clients cannot simultaneously transmit and train, therefore $\mathcal{T}_s^{(i)}+\mathcal{L}_s^{(i)} \leq 1$.


The battery charging event of each user $i$ at slot $s$, denoted $\mathcal{C}_s^{(i)}$, follows a Bernoulli distribution with charging probability $\delta = \text{Pr}[\mathcal{C}_s^{(i)}=1]$.
The battery level of user $i$ evolves as
\begin{equation}\label{eq:battery-evolution}
    E_{s+1}^{(i)} = \max\!\big(E_s^{(i)} - a_s^{(i)},\; 0\big) + \ind_{\{\mathcal{C}_s^{(i)}=1\}},
\end{equation}
where $a_s^{(i)} \in \{0, 1\}$ denotes the per-slot energy cost of client $i$'s action at slot $s$: $a_s^{(i)} = 0$ when idle, and $a_s^{(i)} = 1$ when transmitting or executing a single slot of local training. A full local training session therefore consumes $B$ battery units across $B$ consecutive slots, and a transmission event consumes one battery unit in a single slot.

A client may undertake an action only if it has sufficient energy to complete it; if an action requires more energy than currently available, the action is declined. For example, local training can only be initiated when $E_s^{(i)} \geq B$.


\section{Battery-aware EHFL with Pipelined Cyclic Participation}\label{sect:core}

The proposed algorithm consists of four components, described in the following subsections: group formation, local training, intra-group aggregation, and pipelined cyclic scheduling across groups. The pseudo-algorithm of PipeCycle is detailed in Alg.~\ref{alg:PipeCycle} in Appendix \ref{appdx:pseudo-alg}.

\subsection{Group Formation (Longest-Idle-First)}\label{subsec:group-formation}

Every user harvests one unit of battery with probability $\delta$ at each slot, following the harvest-store-use approach \cite{sudevalayam11}. Before the collaborative learning process, the server assigns users to at most $G$ groups, where $G$ denotes the maximum number of groups allowed to coexist in the pipeline.

At slot $s$, a new group is formed from clients eligible to launch local training. A user $i$ is eligible if it satisfies the following two conditions simultaneously:
\begin{enumerate}
    \item \textbf{Sufficient battery:} $i$ has enough energy to complete a local training session and a subsequent transmission, requiring $E_s^{(i)} \geq B+1$.
    \item \textbf{Idle status:} $i$ is neither transmitting nor training, i.e., $\mathcal{T}_s^{(i)} = \mathcal{L}_s^{(i)} = 0$.
\end{enumerate}

Each group's size is upper-bounded by a threshold $n_G$. When the number of eligible candidates exceeds $n_G$, the server selects the top $n_G$ clients in descending order of idle time $t^{(i)} = s - s^{(i)}_\text{last}$, where $s^{(i)}_\text{last}$ denotes the latest slot at which client $i$ joined a group. This longest-idle-first rule promotes balanced participation across clients and prevents the same subset from monopolizing the pipeline.

\subsection{Local Training}\label{subsec:local-training}

Each group member who has begun local training performs one mini-batch \ac{sgd} step per slot over $B$ slots. With a mini-batch $\xi_s^{(i)}\subset \mathcal{D}^{(i)}$ drawn from its local dataset, client $i$'s local model is updated as
\begin{align}\label{eq:local-SGD}
    \mathbf{x}_{s+1}^{(i)} = \mathbf{x}_{s}^{(i)} - \gamma\, g_i \big(\mathbf{x}_{s}^{(i)};\xi_s^{(i)}\big),
\end{align}
where $\gamma$ is the learning rate and $g_i(\cdot)$ denotes the stochastic gradient. Over $B$ slots of training, the accumulated local update at slot $s$ is
\begin{equation}\label{eq:local-delta}
    \Delta_s^{(i)} = \mathbf{x}_s^{(i)} - \mathbf{x}_{s-B}^{(i)} = -\gamma \sum_{s'=s-B}^{s-1} g_i\big(\mathbf{x}_{s'}^{(i)};\xi_{s'}^{(i)}\big),
\end{equation}
which client $i$ then holds as a pending update until intra-group aggregation occurs.

\subsection{Intra-Group Aggregation}\label{subsec:intra-group-agg}

\begin{figure}[t]
    \centering
    \includegraphics[width=\columnwidth]{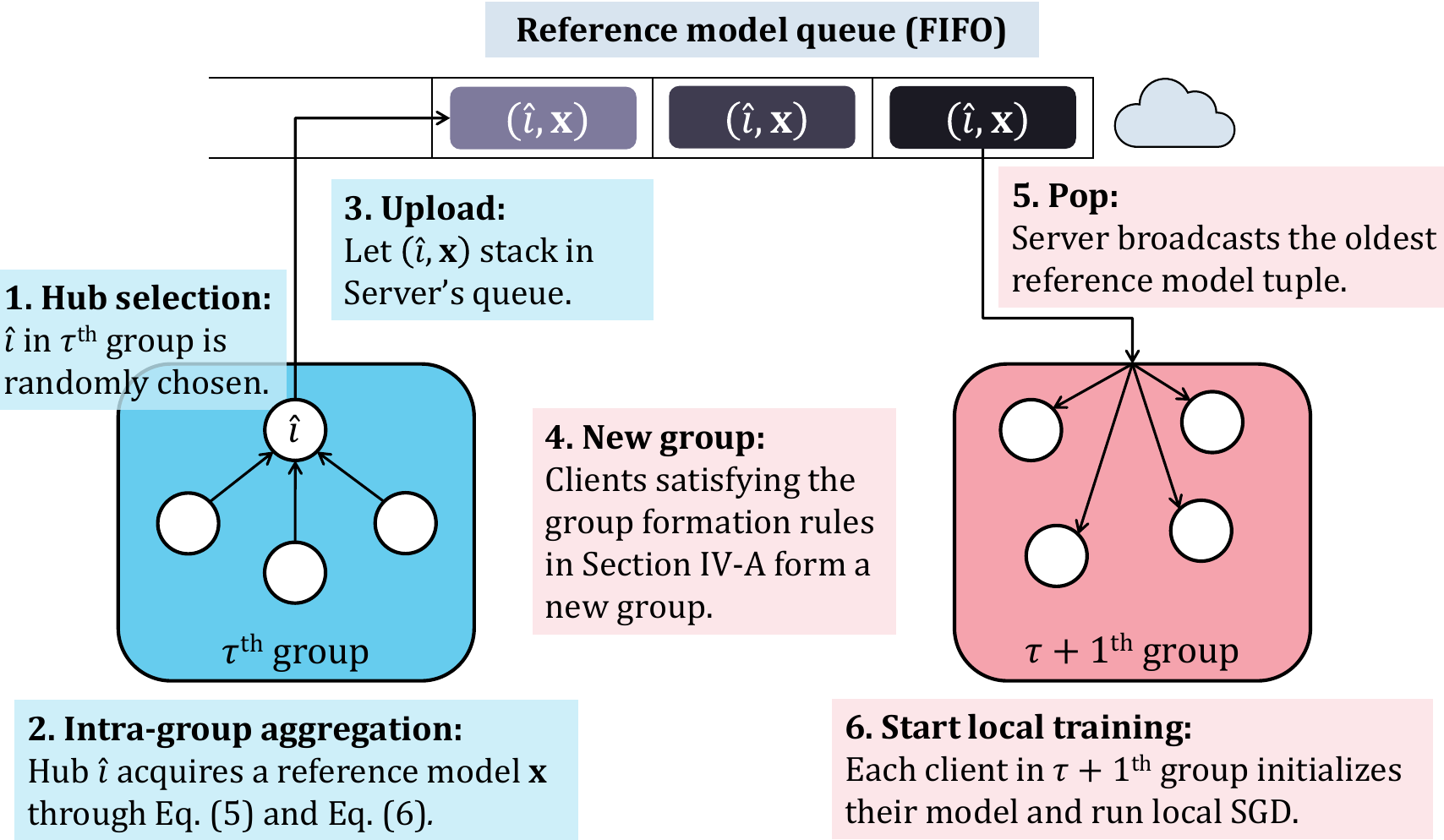}
    \caption{Flow of reference models ($\mathbf{x}$) updated in each group.}
    \label{fig:hub-queue}
\end{figure}

After all members of group $\mathbb{G}_\tau$ (i.e., $\tau$-th group) complete their $B$-slot local training, a temporary hub $\hat{\imath} \in \mathbb{G}_\tau$, shown in Fig.~\ref{fig:hub-queue}, is randomly chosen for intra-group aggregation. All non-hub members transmit their local updates $\Delta^{(i)}$ to $\hat{\imath}$, and the hub computes the aggregated update
\begin{equation}
    \Delta_{\mathbb{G}_\tau} = \sum_{i \in \mathbb{G}_\tau} \frac{n_i}{\sum_{j \in \mathbb{G}_\tau} n_j}\, \Delta^{(i)},
\end{equation}
where $n_i$ denotes the number of samples used by client $i$ during local training. The hub then constructs the group's reference model
\begin{equation}
    \mathbf{x}_{\mathbb{G}_\tau} = \bar{\mathbf{x}}^{(\hat{\imath})} + \Delta_{\mathbb{G}_\tau},
\end{equation}
where $\bar{\mathbf{x}}^{(\hat{\imath})}$ is the initial reference model that the hub received when $\mathbb{G}_\tau$ has just been created.

The server maintains a hub queue that stores pairs of a hub's index and its reference model in first-in-first-out (FIFO) order. When a new group is formed, the server delivers the oldest reference model from the hub queue to the new group as its initial reference model. If the hub queue is empty, the server instead delivers the current global model.

\subsection{Pipelined Cyclic Structure}\label{subsec:pipeline}

\begin{figure}[t]
    \centering
    \includegraphics[width=\columnwidth]{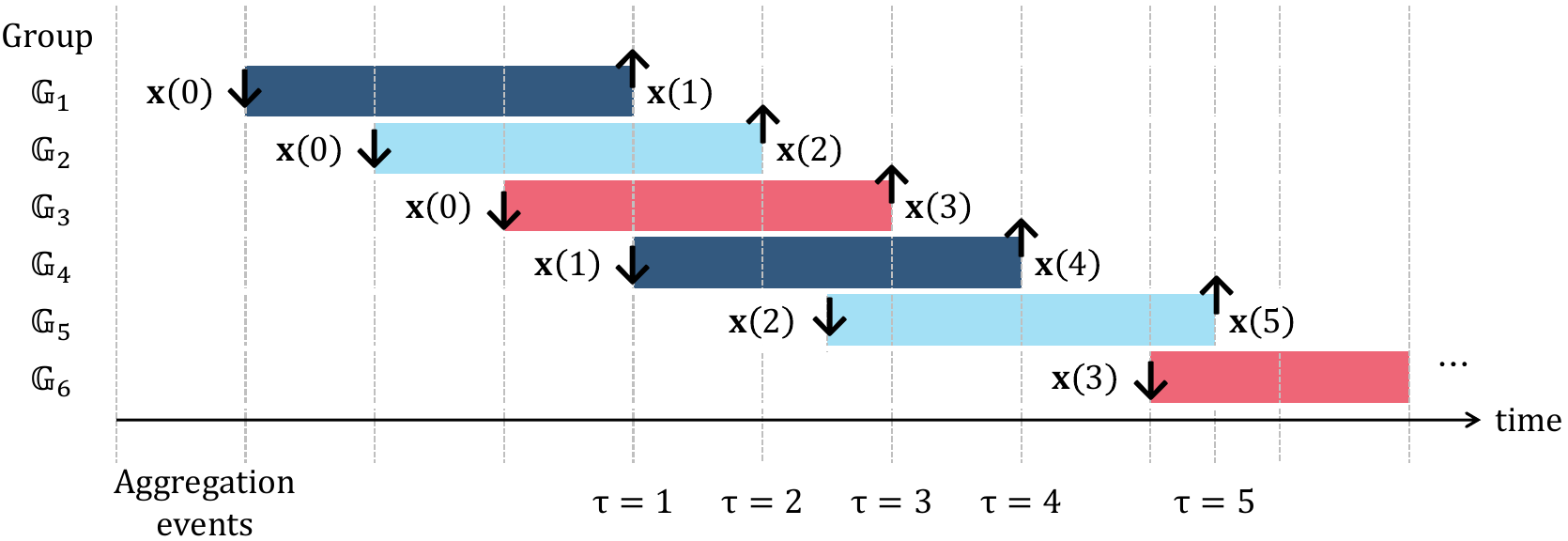}
    \caption{Aggregated model transmission in groups operating PipeCycle. Downward arrows refer to broadcasting reference model from the server's hub queue, whereas upward arrows symbolize uploading reference model from each group after intra-group aggregation.}
    \label{fig:pipelined-cyclic-scheduling}
\end{figure}

Fig.~\ref{fig:pipelined-cyclic-scheduling} illustrates an example of PipeCycle operation with $G=3$, where at most three groups coexist at any given time. We index aggregation events (i.e., moments at which a group completes intra-group aggregation) by $\tau = 1, 2, 3, \ldots$ in chronological order. Let $\mathbf{x}_\tau$ denote the aggregated model produced at event $\tau$, and let $\mathbb{G}_\tau$ denote the group that produces $\mathbf{x}_\tau$. In Fig.~\ref{fig:pipelined-cyclic-scheduling}, the notation $\mathbf{x}(\tau)$ is exceptionally used for better readability.

Each group remains active for $B+1$ slots and is dismissed once its hub uploads the aggregated model to the server. Each color in Fig.~\ref{fig:pipelined-cyclic-scheduling} indicates an ancestry chain: for instance, $\mathbb{G}_4$ adopts $\mathbf{x}_1$, which was produced by $\mathbb{G}_1$, as its initial reference model. Hence $\mathbb{G}_4$ and $\mathbb{G}_1$ belong to the same ancestry chain. Two sequential groups (e.g., $\mathbb{G}_4$ and $\mathbb{G}_5$) may therefore inherit different initial reference models, which can temporarily induce client drift across groups. Bounding this drift is crucial to the convergence analysis in Section \ref{sect:analysis}.


\section{Convergence Analysis}\label{sect:analysis}
In this section, we provide the convergence bound for PipeCycle. 
We adopt the following assumptions that are standard in the optimization literature for establishing convergence.
\begin{assumption}{(Smoothness of local loss functions)}\label{assmp:lipschitz}  The local objective functions of the clients, $f_1(\mathbf{x}), \cdots, f_N(\mathbf{x})$, are all $L$-smooth. That is,
    \[
        \| \nabla f_i(\mathbf{x}) - \nabla f_i(\mathbf{y}) \| \leq L\|\mathbf{x}-\mathbf{y}\| 
    \]
for all $\mathbf{x},\mathbf{y} \in \mathbb{R}^d,\ i\in[N]$.
\end{assumption}

\begin{assumption}{(Unbiased stochastic gradient with bounded variance.)}  \label{assmp:sg}
The expected value of the stochastic gradient $g_i$ is equal to the true (full) gradient of the objective function $f_i$. Additionally, the variance of the stochastic gradient is uniformly bounded by a constant $\sigma^2$. That is, 
    \[
        \mathbb{E}[g_i(\mathbf{x}) | \mathbf{x}] = \nabla f_i(\mathbf{x}) \ \text{and}\ 
        \mathbb{E}\left[ \| g_i(\mathbf{x}) - \nabla f_i(\mathbf{x}) \|^2 | \mathbf{x} \right] \leq \sigma^2
    \]
for all $\mathbf{x}\in \mathbb{R}^d,\ i\in[N]$.
\end{assumption}

Below we present the assumption concerning the clusters $\mathbb{G}_1, \cdots, \mathbb{G}_T$, where $\mathbb{G}_\tau\subset[N]$. Assumption \ref{assmp:group-heterogeneity} generalizes standard data heterogeneity assumptions to the two-level (intra-group and inter-group) structure induced by cyclic participation.
\begin{assumption}{(Intra-group and inter-group data heterogeneity)}\label{assmp:group-heterogeneity}
There exist constants $\tilde{\alpha}, \tilde{\beta} \geq 0$, such that
    \begin{gather*}
        \Big\|\nabla f_i(\mathbf{x}) - \sum_{j\in\mathbb{G}_\tau} \nabla f_j(\mathbf{x}) \Big\| \leq \tilde{\alpha} \text{ and}\\
        \Big\| \sum_{j\in\mathbb{G}_\tau} \nabla f_j(\mathbf{x}) - \nabla f(\mathbf{x}) \Big\| \leq \tilde{\beta}\ ,
    \end{gather*}
for all $\mathbf{x}\in\mathbb{R}^d$, $\tau\in\{1,\cdots,T\}$,  $i,j\in[N]$.
\end{assumption}

\begin{figure}[!t]
\centering
\subfloat[Finding the virtual global model at $\tau=5$.]{
    \includegraphics[width=\columnwidth]{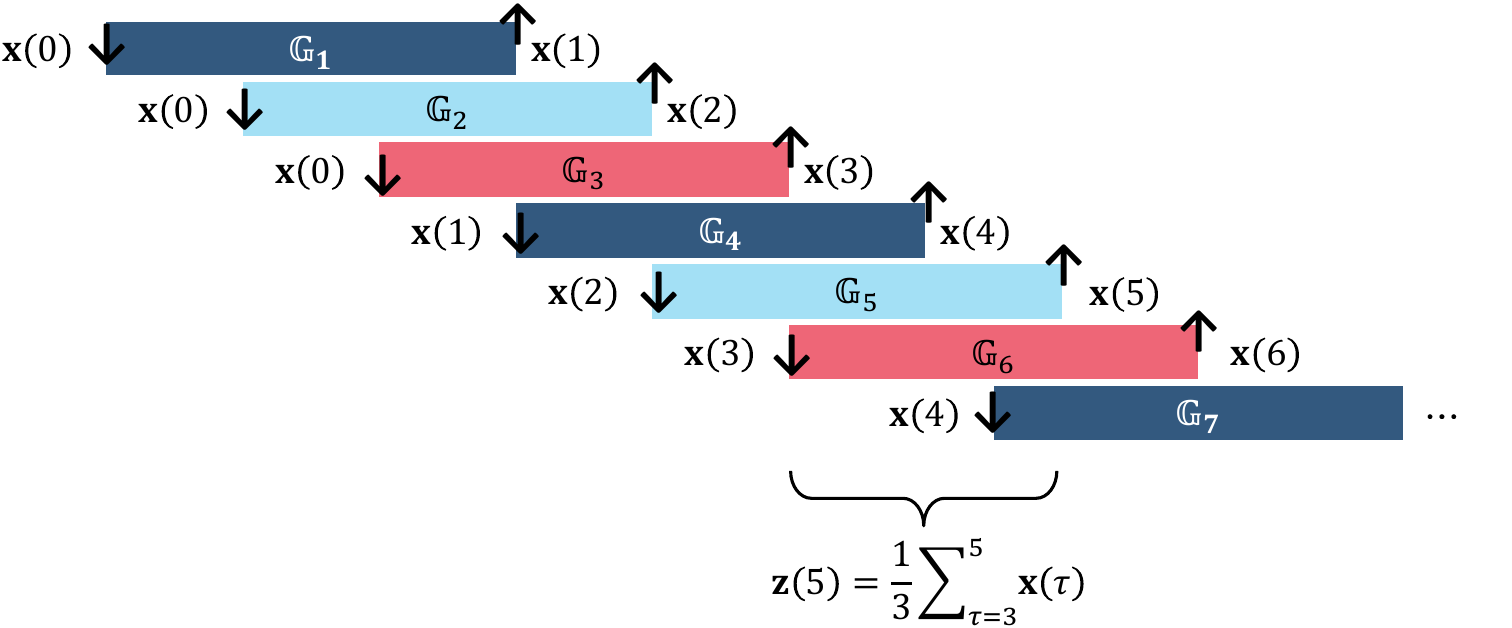}
    \label{fig:analysis-z}
}
\\
\subfloat[The ancestry chain of Group $\mathbb{G}_7$ from Fig.~\ref{fig:analysis-z}.]{
    \includegraphics[width=\columnwidth]{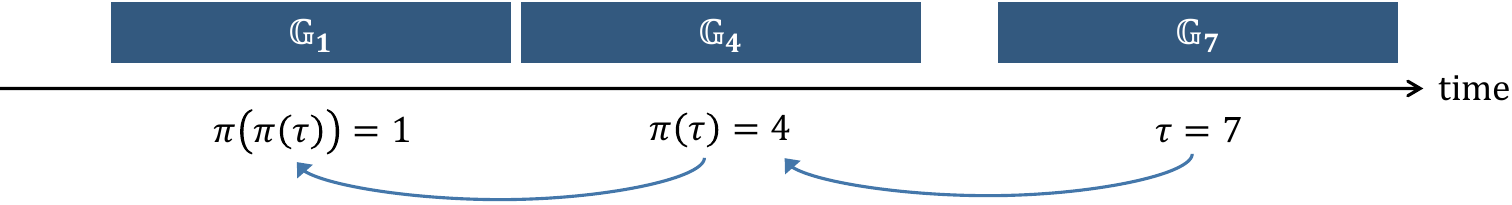}
    \label{fig:analysis-ancestry}
}
\caption{Illustration of PipeCycle's operation considering convergence analysis.}
\label{fig:analysis}
\end{figure}

Adopting an approach similar to that of \cite{mishchenko22}, we define the virtual global model $\mathbf{z}_\tau$ for analyses (see Fig.~\ref{fig:analysis-z}), which is a recursively defined global model. That is,
\begin{align}\label{def:virtual_global_model}
    \mathbf{z}_\tau = \mathbf{z}_{\tau-1}
    +\Delta_{\mathbb{G}_{\tau-1}} ,
\end{align}
where $\mathbf{z}_0 = \mathbf{x}_0$.
Note that $\mathbf{z}_\tau$ appears only in the theoretical analysis, since PipeCycle is semi-asynchronous and clients do not have the same model simultaneously.

The parameter $G$ denotes the maximum number of concurrent groups in the pipeline, as introduced in Section~\ref{subsec:pipeline}. The maximum depth of ancestry chains, $d_{\max}$, characterizes the worst-case staleness of a reference model propagating through the pipeline.
We call $\mathbb{G}_{\pi(\tau)}$ a direct ``ancestor'' of $\mathbb{G}_{\tau}$; in Fig.~\ref{fig:analysis-ancestry}, which depicts only one ancestry chain among the three from Fig.~\ref{fig:analysis-z}, we have $\pi(7)=4$ and $\pi(4)=1$.
When $\tau$-th group inherits its initial reference model $\mathbf{x}_{\pi(\tau)}$ from a predecessor (i.e., $\pi(\tau)$-th group) that has itself inherited from an earlier predecessor (i.e., $\pi(\pi(\tau))$-th group), and so on, the length of such chains of inherited models is bounded by
\begin{align}\label{eq:dmax}
    d_{\max} = \left\lfloor \frac{S_{\mathrm{tot}}}{B} \right\rfloor .
\end{align}
This finite-horizon bound comes from the fact that each group remains active for at least $B$ slots and no two aggregation events on the same ancestry chain can occur within the same $B$-slot window. The bounded staleness is a structural property of cyclic participation, and it avoids the exponential factors that typically appear in convergence analyses of asynchronous \ac{fl}~\cite{mishchenko22}. Throughout this section, the $\tau$-th aggregation event refers to the moment at which a group completes its intra-group aggregation and is dismissed. We denote by $\mathbb{G}_\tau \subseteq [N]$ the group dismissed at this event, by $\mathbf{x}_\tau$ the aggregated reference model that $\mathbb{G}_\tau$ produces immediately before dismissal, and by $T$ the total number of such events over the horizon $S_{\mathrm{tot}}$. 
With these structural quantities in place, we now state the main convergence result for PipeCycle. The theorem bounds the minimum expected squared gradient norm of the virtual global model along the trajectory.

\begin{theorem}{(Bounded stationarity gap of the global model) \label{thm:0}}Under Assumptions \ref{assmp:lipschitz}, \ref{assmp:sg}, and \ref{assmp:group-heterogeneity} hold, for a learning rate $\gamma<\frac{1}{4B L(d_{\max}^2+1)(G-1)}$, the minimum expected squared gradient norm across the trajectory is bounded by
    \begin{align}\label{ineq:thm:0}
        \min_{\tau}\mathbb{E}[\|\nabla f(\mathbf{z}_\tau)\|^2]\leq \frac{f(\mathbf{x}_0) - f^*}{\zeta T} + \frac{\Lambda^2}{\zeta} ,
    \end{align}
where\[
    \zeta=\frac{B\gamma}{2} -2B^2\gamma^2L -2B^2\gamma^2 (d_{\max}^2+1)L(G-1)^2
\] and
\begin{align*}
   \Lambda^2 &= L\big((d_{\max}+1)(G-1)+1\big) \cdot\\
   &\quad(B\gamma^2\sigma^2\omega^2
        + 4 L^2B^2\gamma^2
        + 4 B^2\gamma^2 (\tilde{\alpha}+\tilde{\beta})^2)
        +\frac{\tilde{\beta}^2B\gamma}{2} .
\end{align*}
\end{theorem}
\begin{proof}
    Provided in Appendix~\ref{appdx:proof-thm}.
\end{proof}

The total number of aggregation events ($T$) depends on the battery charging probability $\delta$ and the maximum number of coexisting groups $G$ in operation. Hence, we provide a system-level interpretation in Corollary~\ref{corol:0}.
\begin{corollary}{(System-level bound)\label{corol:0}}
    The convergence bound from Theorem~\ref{thm:0} can be expressed in terms of system parameters by substituting the lower bound on $T$, i.e.,
    \begin{align*}
        \min_{\tau}\mathbb{E}[\|\nabla f(\mathbf{z}_\tau)\|^2]\leq \frac{f(\mathbf{x}_0) - f^*}{S_{\mathrm{tot}}\zeta \cdot \min\left(\frac{N\delta}{n_G(B+1-B\delta)},\frac{G}{B}\right)} + \frac{\Lambda^2}{\zeta}
    \end{align*}
\end{corollary}
\begin{proof}
    Provided in Appendix~\ref{appdx:proof-corol}.
\end{proof}
Larger $G$ helps the transient term decay faster through $T$, but it inflates the asymptotic stationarity gap that the algorithm settles toward. Theoretically, however, the bound implies that for very long horizons where the transient vanishes, smaller $G$ achieves a tighter floor. This trade-off suggests that there is a pipeline capacity that strikes the best balance between the two competing effects. If we adopt the rule $n_G=\lfloor \frac{N}{2.5G}\rfloor$, which couples the group size to $G$, the asymptotic stationarity gap in Theorem~\ref{thm:0} becomes a function over $G$ alone, and we can derive a closed-form expression for its minimizer. This rule is used in our experimental setup in Section~\ref{sect:exp}.

\noindent{\textbf{Remark.}} \emph{(Optimal $G$)}\label{optimal_G} When the number of each group's members is bounded by $n_G=\lfloor \frac{N}{2.5G}\rfloor$, the transient's denominator of (\ref{ineq:thm:0}) becomes a function over $G$. The tightest possible bound of Theorem~\ref{thm:0} is attained at an optimal $G$,
\begin{align}\label{eq:G-star}
    G^\star = \frac{2}{3} + \frac{1}{3}\sqrt{1+\frac{3(1-4B\gamma L)}{4B\gamma(d_{\max}^2+1)L}} ,
\end{align}
which is valid for any $\delta\in(0,1]$.
\begin{proof}
    Provided in Appendix~\ref{appdx:proof-remark}.
\end{proof}
$G^\star$ from (\ref{eq:G-star}) minimizes the analytical upper bound under the adopted group-size rule. It is not a universal performance optimum across all finite-horizon settings and data distributions.


\section{Experimental Results}\label{sect:exp}

\subsection{Experimental Setup}

We evaluate PipeCycle on image classification using the CIFAR-10 dataset, partitioned across $N = 100$ clients, with each client holding 300 training samples. We use a neural network consisting of two convolutional layers, one max-pooling layer, and three fully connected layers. When trained in isolation on a single client's local data, this model achieves only about 10\% test accuracy, which is equivalent to random guessing across the ten classes and necessitates collaboration with the others.

We consider both IID and Non-IID data distribution, where data heterogeneity is generated using a Dirichlet distribution with concentration parameter $\alpha = 0.1$, producing severe label skew across clients. Each client performs mini-batch \ac{sgd} with a batch size of 15, learning rate $\gamma = 0.05$, and cross-entropy loss. A single local training session occupies $B = 20$ successive slots and consumes 20 battery units, and each uplink transmission consumes one battery unit. For clustering-based strategies, we test $G \in \{2, 5, 10\}$.
The maximum group size is set to $n_G = \lfloor \frac{N}{2.5G} \rfloor$, which gives $n_G \in \{20, 8, 4\}$ for the respective $G$ values at $N = 100$.

The battery charging probability varies from extremely energy-constrained to always-supplied, ranging within $\delta \in \{0.1, 0.3, 0.5, 1.0\}$. The maximum battery capacity is set to $E_{\max} = 10^6$, and all clients begin with an empty battery, $E_0^{(i)} = 0$. The total training horizon is $S_{\mathrm{tot}} = 15{,}000$ slots for IID experiments and $S_{\mathrm{tot}} = 30{,}000$ slots for non-IID experiments, reflecting the slower convergence under heterogeneous data. We evaluate every 150 slots on a held-out test set of 1{,}000 samples drawn uniformly from the CIFAR-10 official test set. To prevent unnecessary computation when a training curve has saturated, we apply early stopping whenever the test accuracy varies by less than 1\% across 10 checkpoints in a row. Further details on the experiment settings are available in Appendix \ref{appdx:exp-settings}.

\subsection{Baselines and Adjustments}

We benchmark PipeCycle against several established \acf{fl} schemes, categorized based on their consideration of \ac{ehfl}.
\begin{itemize}
    \item \Ac{fl} without \ac{eh}: These schemes do not have system models that account for \ac{ehfl}. We implemented them under the same battery charging conditions as PipeCycle to ensure a fair comparison.
    \begin{itemize}
        \item FedAvg \cite{mcmahan17:fedavg}
        \item CyCP + SGD \cite{cho23} 
        \item MIFA \cite{gu21:mifa}
    \end{itemize}
    \item \Ac{ehfl}: These schemes are designed to operate in \ac{ehfl} environments.
    \begin{itemize}
        \item FedSeq \cite{chen23:FedSeq} 
        \item FLDA \cite{valentedasilva25:FLDA}
        \item FedBacys \cite{jeong25:FedBacys}
    \end{itemize}
\end{itemize}
For all schemes, including the baselines, we treat all $N$ users as potential participants in every epoch. However, users may fail to participate in aggregation due to insufficient battery power or an unprepared local update at the time of aggregation.

In addition to time slots $s=\{1,\cdots,S_{\mathrm{tot}}\}$, two time units, \emph{epoch} and \emph{group-round}, are introduced in the baseline algorithms. An epoch represents a complete communication round where the server aggregates local model updates to refine the global model. In frameworks with clustered scheduling policies, a group-round $R$ corresponds to the duration of a group's activation window. We denote $S$ as the number of slots per epoch. For example, when $S=30$ and the clients are divided into $G=3$ groups, the aggregation process within each group takes $R=\lfloor \frac{S}{G} \rfloor = 10$ slots sequentially.

We evaluate the effectiveness of each scheme using two key metrics: total energy consumption and time-to-performance. Specifically, we count the cumulative battery units consumed by the entire system throughout the learning process. This metric provides insight into the energy efficiency of each algorithm. We also compare how quickly each method achieves a desired performance level by plotting the saturation curves of test accuracy as a function of the number of epochs.

In the experiments, we adjust the benchmark \ac{fl} frameworks due to the disparity in local training duration within the algorithm as follows.   

    \noindent\textbf{Adjustment to FedAvg~\cite{mcmahan17:fedavg}: }
    Vanilla FedAvg does not account for potential energy scarcity. We thereby implant energy harvesting protocol to FedAvg and let the clients always prioritize local training whenever available. All users train locally \emph{in a greedy manner}, i.e., as soon as $E_s^{(i)}\geq B$, they launch local training while using their freshest reference model. If a user has an update that hasn't been sent yet, it attempts uplink transmission as soon as $E_s^{(i)}\geq 1$.

    \noindent\textbf{Adjustment to CyCP+SGD~\cite{cho23}: }
    The original CyCP algorithm operates only when clients are randomly chosen at the beginning of each group round and finish local training within the unit time slot. However, we consider the case where there is always $B>1$ slots of time gap to participate in the global aggregation. 
    \begin{itemize}
    	\item All users perform greedy local training as they do in adjusted FedAvg.
    	\item At each slot $s$ where $s\equiv R-1$ (mod $R$), among the users, those who have a pending local update and at least one battery unit stored are chosen for the current group.
    	\item If this group is an empty set, the process is skipped, and we try the next slot where $s\equiv R-1$ (mod $R$).
    	\item The maximum number of group members is $\lfloor\frac{N}{G}\rfloor$. If there are more than $\lfloor\frac{N}{G}\rfloor$ users that satisfy the group-joining criteria, we randomly pick $\lfloor\frac{N}{G}\rfloor$ users as the group's members. In contrast, the residual users will participate in the next group. Once the group is confirmed, we randomly select the star node that collects the local updates and aggregates them to update the intermediate global model.
    \end{itemize} 
    
    \noindent\textbf{Adjustment to MIFA~\cite{gu21:mifa}: }
    The algorithm assumes the active clients in each global round are arbitrarily chosen, following an IID Bernoulli distribution with the participation probability $p_i$ for each client $i$. The users possessing training data with rare class labels have small participation probabilities. The adjusted MIFA based on longer local training sessions follows the rule:
    \begin{itemize}
    	\item At time slot $s\equiv 0$ (mod $S$), the server broadcasts $\mathbf{x}_t$ to all clients, where $t$ denotes the current epoch.
    	\item When  $s\equiv S-B-1$ (mod $S$), the users satisfying the learning criteria form a subset: these users in the subset begin local model training and return $\Delta_t^{(i)}$.
    	\item When  $s\equiv S-1$ (mod $S$), users who have unsent $\Delta_t^{(i)}$ transmit $\Delta_t^{(i)}$ to the server. The server aggregates the received local updates and feeds their weighted sum to renew the global model.
    \end{itemize}

    This algorithm alternates \ac{fl} and Federated Distillation (FD), which have negligible difference in energy consumption for local computation. Instead, the transmission costs for \ac{fl} and FD have a substantial gap, as these costs are proportional to the size of the uploaded local update.
    In our system model, the size of the local update is equivalent to that of the \ac{fl} case in FLDA. Thus, we can calculate the energy consumption for transmitting one FD local update by multiplying the ratio of the dimensions of the uploading local updates. Since each \ac{fl} update consumes 1 battery unit, each FD update consumes $1\cdot d_\text{FD}/d_\text{FL}$ battery unit, where $d_\text{FD}$ and $d_\text{FL}$ refer to the output dimension and the entire model parameters' dimension, respectively.

\subsection{Results}

\begin{table}[tb]
\caption{%
    Global model test accuracy (\%) achieved within an energy budget of
    $5\times10^{5}$ units under IID data distribution.
    \label{table:iid}}
\centering
\renewcommand{\arraystretch}{1.1}
\begin{tabular}{|l|c||c|c|c|c|}
\hline
\multirow{2}{*}{Strategy} & \multirow{2}{*}{$G$}
    & \multicolumn{4}{c|}{Battery charging probability $\delta$} \\
\cline{3-6}
 & & 0.1 & 0.3 & 0.5 & 1.0 \\
\hline
FedAvg   & $-$ & 65.4 & 64.8 & 61.4 & 50.4 \\
\hline
MIFA     & $-$ & 62.2 & 62.7 & 60.5 & 55.9 \\
\hline
FLDA     & $-$ & 64.6 & 64.4 & 63.2 & 56.6 \\
\hline
\multirow{3}{*}{CyCP+SGD}
    &  2 & 65.3 & 65.5 & 63.4 & 57.2 \\
\cline{2-6}
    &  5 & 62.7 & 65.0 & 67.1 & 63.9 \\
\cline{2-6}
    & 10 & 43.4 & 50.3 & 46.3 & 40.5 \\
\hline
\multirow{3}{*}{FedSeq}
    &  2 & 64.9 & 63.9 & 64.1 & 64.5 \\
\cline{2-6}
    &  5 & 59.9 & 65.3 & 52.5 & 51.6 \\
\cline{2-6}
    & 10 & 36.7 & 44.1 & 27.6 & 25.5 \\
\hline
\multirow{3}{*}{FedBacys}
    &  2 & 38.4 & 45.0 & 57.6 & 61.8 \\
\cline{2-6}
    &  5 & 17.2 & 14.8 & 19.6 & 58.8 \\
\cline{2-6}
    & 10 & 18.3 & 12.5 & 19.0 & 17.4 \\
\hline
\multirow{3}{*}{\textbf{PipeCycle}}
    &  2 & 68.5 & 69.0 & 69.4 & 69.2 \\
\cline{2-6}
    &  5 & 67.3 & \textbf{70.2} & 69.9 & 69.6 \\
\cline{2-6}
    & 10 & \textbf{66.5} & 69.1 & \textbf{72.0} & \textbf{71.9} \\
\hline
\end{tabular}
\end{table}

\begin{table}[tb]
\caption{%
    Global model test accuracy (\%) achieved within an energy budget of
    $5\times10^{5}$ units under \textbf{non-IID} data distribution
    (Dirichlet, $\alpha=0.1$).
    \label{table:noniid}}
\centering
\renewcommand{\arraystretch}{1.1}
\begin{tabular}{|l|c||c|c|c|c|}
\hline
\multirow{2}{*}{Strategy} & \multirow{2}{*}{$G$}
    & \multicolumn{4}{c|}{Battery charging probability $\delta$} \\
\cline{3-6}
 & & 0.1 & 0.3 & 0.5 & 1.0 \\
\hline
FedAvg   & $-$ & 62.7 & 58.7 & 51.1 & 41.9 \\
\hline
MIFA     & $-$ & 39.1 & 43.0 & 19.6 & 46.7 \\
\hline
FLDA     & $-$ & 63.4 & 59.9 & 51.6 & 43.5 \\
\hline
\multirow{3}{*}{CyCP+SGD}
    &  2 & 60.0 & 55.4 & 44.8 & 31.1 \\
\cline{2-6}
    &  5 & 10.3 & 12.6 & 16.2 & 50.3 \\
\cline{2-6}
    & 10 & 10.3 & 10.1 & 10.0 & 10.8 \\
\hline
\multirow{3}{*}{FedSeq}
    &  2 & \textbf{65.6} & 63.2 & 60.4 & 60.1 \\
\cline{2-6}
    &  5 & 10.2 & 12.2 & 12.2 & 12.2 \\
\cline{2-6}
    & 10 & 10.3 & 10.3 & 10.3 & 10.3 \\
\hline
\multirow{3}{*}{FedBacys}
    &  2 & 10.0 & 10.0 & 10.6 & 10.2 \\
\cline{2-6}
    &  5 & 10.2 & 10.1 & 10.0 & 10.5 \\
\cline{2-6}
    & 10 & 10.2 & 10.1 & 10.0 & 10.1 \\
\hline
\multirow{3}{*}{\textbf{PipeCycle}}
    &  2 & 65.0 & \textbf{66.1} & \textbf{63.7} & \textbf{64.5} \\
\cline{2-6}
    &  5 & 61.1 & 62.3 & 61.8 & 61.5 \\
\cline{2-6}
    & 10 & \textbf{65.3} & 55.7 & 52.7 & 45.8 \\
\hline
\end{tabular}
\end{table}

Table~\ref{table:iid} shows the achieved test accuracy given a fixed energy budget of $5\times10^5$ units, where the bold figures indicate the column-wise best results. The number of groups is not considered in FedAvg, MIFA, and FLDA, since they do not have clustering policies for participant selection. Regardless of the battery charging probability ($\delta$), PipeCycle consistently reaches the highest global model test accuracy. Furthermore, among the grouping-based approaches, PipeCycle is the only method that does not experience drops when $G$ increases.

For non-IID experiments, we apply an additional rule to the hub queue to prevent stale reference models: when the queue grows beyond a threshold due to infrequent group formation under extreme energy scarcity, the oldest reference model is discarded.\footnote{This is an engineering safeguard rather than a core algorithmic component and is not used in the IID experiments.} 
Table~\ref{table:noniid} summarizes the test accuracy achieved under non-IID data distribution with the same fixed energy budget. Differentiated from the IID cases explained in Table~\ref{table:iid}, grouping-based methods break under non-IID with larger $G$. In contrast, PipeCycle remains functional up to $G=5$ for $\delta=\{0.3, 0.5, 1.0\}$. However, it also experiences performance degradation in $G=10$, which implies that having at most 10 concurrent groups under harsh non-IID may fragment the users and lead to larger client drift.

The closed-form $G^\star$ in Corollary~\ref{corol:0} is independent of $\delta$, since the total number of aggregation events $T$ scales linearly with $G$ in both cases of the minimum in Corollary~\ref{corol:0}. Battery charging probabilities change the absolute speed at which aggregation events accumulate: when batteries are rarely charged, $T$ grows slowly with the training horizon, leading to slower decay of the transient term $\frac{f(\mathbf{x}_0)-f^*}{\zeta T}$. Within the finite training horizon $S_{\mathrm{tot}}$, the algorithm therefore does not reach the asymptotic floor at low $\delta$, and configurations with smaller $G$, which have a tighter floor, outperform larger $G$ that would only be successful over much longer horizons. Our experimental results are consistent with this finite-horizon behavior: at $\delta = 0.1$ under non-IID data, PipeCycle achieves its best accuracy at $G = 2$, while at higher $\delta$,  the gap between $G$ values narrows.

\begin{figure}[!t]
    \centering
    \includegraphics[width=\columnwidth]{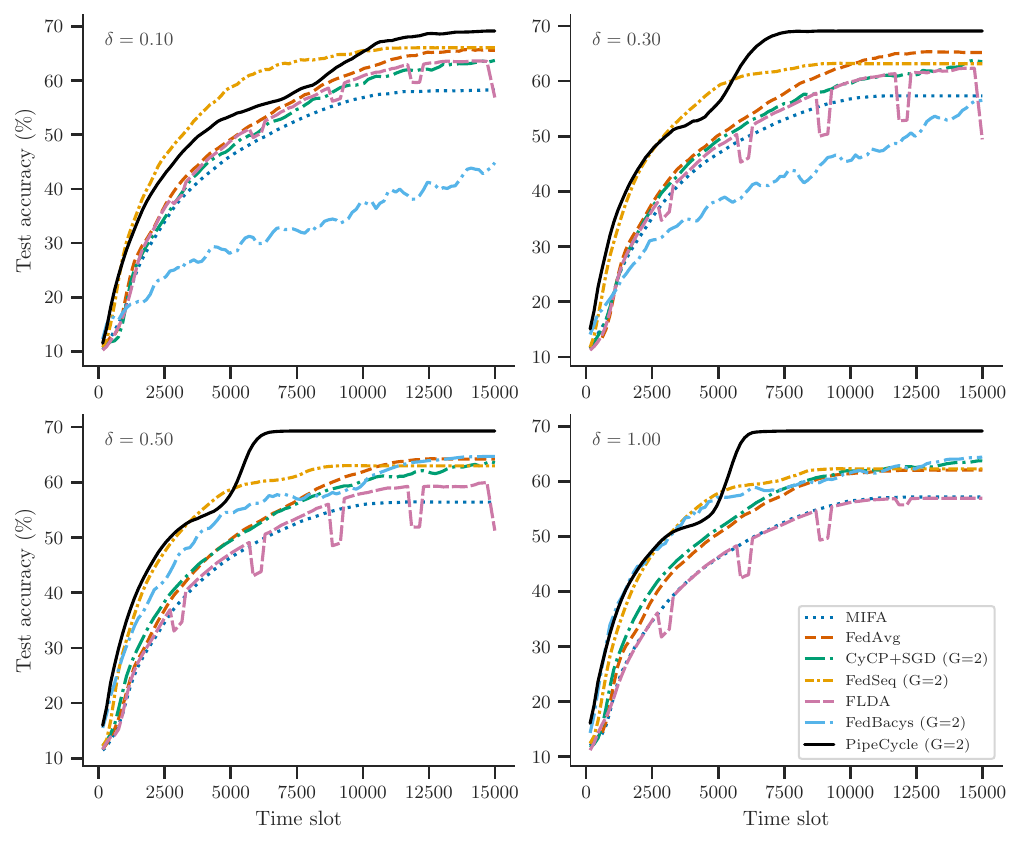}
    \caption{Average test accuracies under IID data distribution from operating different FL algorithms with respect to slots for $G=2$.}
    \label{plot:testacc-iid}
\end{figure}

Fig.~\ref{plot:testacc-iid} shows the arithmetic mean test accuracy across all $N$ clients over the training horizon under IID data, with $G = 2$ for clustering-based methods. PipeCycle reaches the highest final test accuracy across all values of $\delta$. Notably, this holds even at $G = 2$, which is not its strongest configuration: Table~\ref{table:iid} shows that larger $G$ values yield higher final accuracy for PipeCycle, yet even the $G=2$ configuration outperforms every baseline at its own best $G$. PipeCycle and FedSeq exhibit the fastest early-stage rise in accuracy among all baselines: at $\delta \geq 0.3$, PipeCycle reaches 50\% accuracy slightly before FedSeq, while at $\delta = 0.1$ FedSeq edges ahead by a small margin. The remaining baselines lag behind both throughout the early training stage.

\begin{figure}[!t]
    \centering
    \includegraphics[width=\columnwidth]{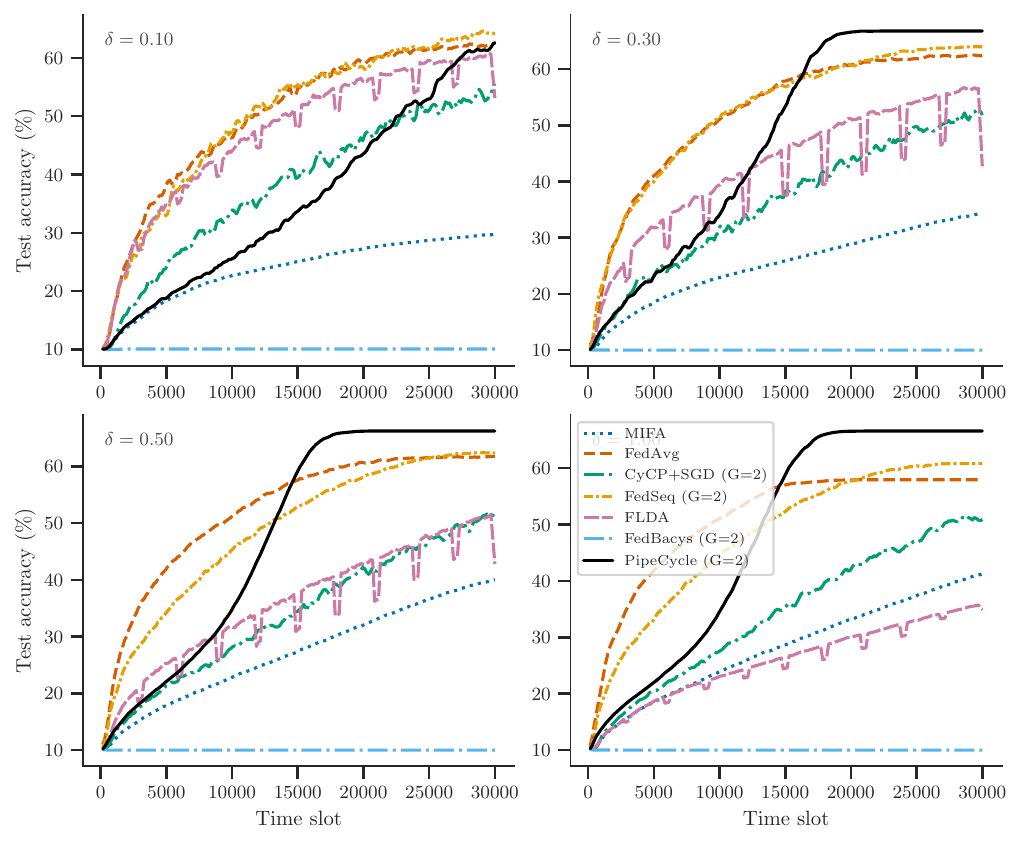}
    \caption{Average test accuracies under Non-IID data distribution from operating different FL algorithms with respect to slots for $G=2$.}
    \label{plot:testacc-noniid}
\end{figure}

Fig.~\ref{plot:testacc-noniid} compares the same approaches under non-IID data with Dirichlet concentration $\alpha = 0.1$. For $\delta \geq 0.3$, PipeCycle achieves the highest final test accuracy across all baselines, maintaining 60\% or higher accuracy at $G = 2$ while clustering-based baselines either degrade with $\delta$ or collapse to chance accuracy at larger $G$ values (see Table~\ref{table:noniid}). The robustness of PipeCycle across $G \in \{2, 5, 10\}$ is particularly noteworthy: under severe label skew, FedSeq, CyCP, and FedBacys all collapse to roughly 10\% accuracy at $G = 10$, while PipeCycle preserves 45\% accuracy or higher across the same range. This indicates that the pipelined relay mechanism mitigates the per-group aggregation quality loss that breaks the alternative cyclic schemes under non-IID conditions.

At the most energy-constrained setting, $\delta = 0.1$, PipeCycle converges more slowly than FedAvg and FLDA, requiring a larger number of slots to reach a comparable test accuracy. This behavior is consistent with the energy-limited regime identified in Section~\ref{sect:analysis}. When battery recharging rarely happens, the total number of aggregation events $T$ no longer scales with the pipeline capacity $G$, which largely weakens the structural advantage of overlapping group rounds. PipeCycle still attains comparable test accuracy at $\delta = 0.1$ (65.0\% at $G = 2$, comparable to FedSeq's 65.6\%), but its convergence speed in slots is no longer favorable in this environment. We note that FedBacys, which shares the cyclic structure but lacks the pipelined relay, fails entirely across all $\delta$ values, suggesting that the relay mechanism is what enables PipeCycle to improve learning performance in the non-IID setting.

\begin{figure}[!t]
    \centering
    \includegraphics[width=\columnwidth]{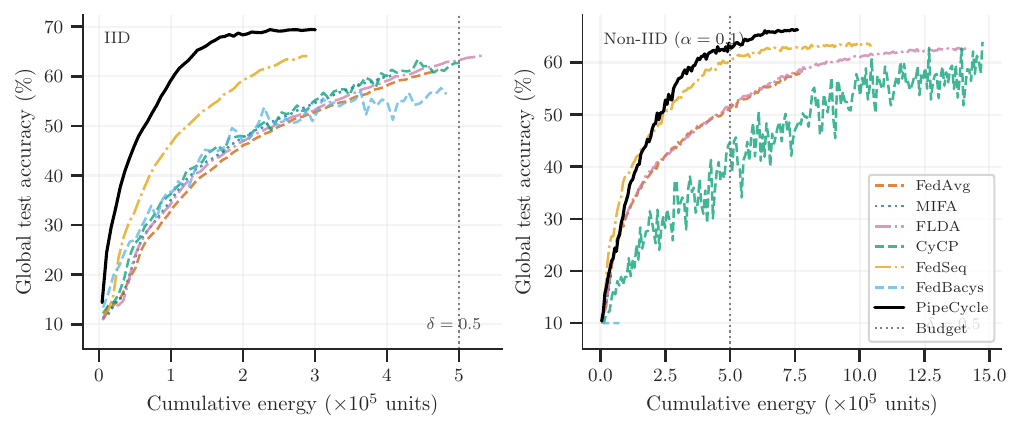}
    \caption{Global test accuracy with respect to cumulative energy consumption ($\delta=0.5$)}
    \label{plot:energy-to-acc}
\end{figure}

Fig.~\ref{plot:energy-to-acc} reports global model test accuracy as a function of cumulative network-wide energy consumption. Our proposed framework dominates the accuracy-energy trade-off across both data distributions. For any fixed energy budget shown, it achieves accuracy at least as high as the best baseline. Notably, for any fixed target accuracy, it spends the smallest cumulative energy consumption. The gap widens noticeably under non-IID, with PipeCycle reaching 65\% global accuracy using roughly half the energy required by the strongest competing baseline. In sum, our proposed framework shows the most practical advantage when energy budget is the binding constraint by offering the most accuracy per unit of network-wide energy spent.

\section{Conclusions}\label{sect:conc}

We introduced PipeCycle, a novel battery-aware framework for \acf{ehfl} that employs pipelined cyclic client participation. By explicitly modeling the substantial time and energy required for local training, PipeCycle organizes clients into pipelined cyclic groups and relays aggregated reference models directly between consecutive groups. This design allows multiple groups to coexist in the pipeline, overlapping the recharging periods of inactive clients with active training in other pipeline stages, and leads to significant improvements in accuracy-energy efficiency compared to existing \ac{ehfl} algorithms.
Our theoretical analysis exposes a finite-horizon staleness bound that avoids the exponential factors typical of asynchronous FL analyses, and yields a closed-form expression for the pipeline capacity $G^\star$ that minimizes the analytical convergence bound under a coupling between group size and the number of concurrent groups.

As future directions, our framework can be extended to develop scheduling strategies that effectively manage users with heterogeneous battery charging probabilities, or to incorporate semantics-aware adaptive participation based on both resource availability and update significance.

\bibliographystyle{ieeetr}  
\bibliography{ref}

{\appendices
\section{Preliminaries}

We define $\pi(\tau)$ as the aggregation event whose output served as the reference model for the group that produced $\mathbf{x}_\tau$, as illustrated in Fig.~\ref{fig:analysis-ancestry}.
In single-pipeline FL, $\pi(\tau)=\tau-1$. Meanwhile, in multiple-pipelined FL as in our framework of which the maximum mumber of coexisting groups is $G$, the lag between two aggregation events in the same pipeline can be greater than $1$, but it is always bounded to $G$. In other words, $\tau-\pi(\tau)\leq G$ for all $\tau$.

Derived from the definition of $\mathbf{z}_\tau$ in (\ref{def:virtual_global_model}), we have
\allowdisplaybreaks{
\begin{gather}\label{expansion_ztau}
    \mathbf{z}_0 = \mathbf{x}_0 , \nonumber \\
    \mathbf{z}_{\tau+1} = \mathbf{z}_\tau + \bar{\Delta}_{\mathbb{G}_{\tau+1}} ,\\
    \mathbf{z}_\tau = \mathbf{x}_0 + \sum_{m=1}^\tau \bar{\Delta}_{\mathbb{G}_m}. \nonumber
\end{gather}}
The actual iterate $\{\mathbf{x_\tau}\}$ follows the relay chain. In other words, $\mathbf{x_\tau}$ is calculated from $\mathbf{x}_{\pi(\tau)}$ and not its previous group's reference model $\mathbf{x}_{\tau-1}$. Thus, we define an ancestry of event epoch $\tau$, denoted by $\mathbb{A}(\tau)=\{\tau, \pi(\tau), \pi(\pi(\tau)),\cdots,0\}$. Then we have a closed-form expression for the actual iterate:
\begin{align*}
    \mathbf{x}_\tau = \mathbf{x}_0 + \sum_{m\in\mathbb{A}(\tau), m\geq 1} \bar{\Delta}_{\mathbb{G}_m} .
\end{align*}
We also define the perturbation $\mathbf{e}_\tau$, which is the error term between the virtual global model and the intra-group model obtained after the local training session,
\begin{align}\label{def:perturbation}
    \mathbf{e}_\tau = \mathbf{z}_\tau - \mathbf{x}_\tau,
\end{align}
which is analogous to the stochastic error term in \cite{mania17:perturbed_iter}.

The weighted average of aggregated update within group $\mathbb{G}_\tau$, denoted as $\bar{\Delta}_{\mathbb{G}_\tau}$, is defined as
\begin{align}\label{def:intra_group_Delta}
    \bar{\Delta}_{\mathbb{G}_\tau} = \sum_{i\in\mathbb{G}_\tau} \underbrace{\frac{|\xi_i|}{\sum_{j\in\mathbb{G}_\tau}|\xi_j|}}_{:=w^{(i)}} \Delta_\tau^{(i)},
\end{align}
where $|\xi_i|$ is the number of training samples comprising the local training batch of client $i$. We set the upper bound of the sum of the squared weights $\omega^2$,
\begin{align}\label{def:omega}
    \omega^2 = \max_{\tau\in(0,T]} \sum_{i\in\mathbb{G}_\tau} (w^{(i)})^2 ,
\end{align}
which is bounded $\omega^2\in[\frac{1}{n_G}, 1]$. In this regard, $\omega^2$ becomes closer to $\frac{1}{n_G}$ when batch sizes are balanced across clients. 

All notations are summarized in Table~\ref{table:nomenclature}.
\begin{table}[tb]
\caption{Nomenclature \label{table:nomenclature}}
\renewcommand{\arraystretch}{1.3}
\centering
\begin{tabular}{c|p{0.7\columnwidth}}
\hline
Notation & \multicolumn{1}{c}{Description} \\ \hline

\multicolumn{2}{c}{\textit{Structural Parameters}} \\ \hline

$B$                & Local training duration (slots) \\
$R$                & Minimum group-formation interval (slots) \\
$G$                & Maximum concurrent groups \\
$n_G$              & Maximum members per group \\
$N$                & Total number of clients \\
$\delta$           & Battery charging probability per slot \\
$E_{\max}$         & Maximum battery capacity \\
$S_{\mathrm{tot}}$ & Total number of time slots \\
$T$                & Total number of aggregation events \\
$d_{\max}$         & Max.\ ancestry chain depth ($\lfloor S_{\mathrm{tot}} / B \rfloor$) \\ \hline

\multicolumn{2}{c}{\textit{Models and Losses}} \\ \hline

$\mathbf{x}^{(i)}$ & Local model of client $i$ \\
$\mathbf{z}$        & Virtual global model (convergence analysis) \\
$f_i(\cdot)$        & Local loss function of client $i$ \\
$f(\cdot)$          & Global loss function \\
$E_s^{(i)}$         & Battery level of client $i$ at slot $s$ \\ \hline

\multicolumn{2}{c}{\textit{Local Training}} \\ \hline

$\xi_s^{(i)}$      & Mini-batch drawn from $\mathcal{D}^{(i)}$ at slot $s$ \\
$|\xi_s^{(i)}|$    & Mini-batch size of client $i$ at slot $s$ \\
$g_i(\mathbf{x};\xi)$ & Stochastic gradient of $f_i$ at $\mathbf{x}$ over batch $\xi$ \\
$\Delta_\tau^{(i)}$ & Local update of client $i$ at aggregation event $\tau$ \\
$\bar{\Delta}_{\mathbb{G}_\tau}$ & Weighted aggregated update of group $\mathbb{G}_\tau$ \\
$w^{(i)}$          & Aggregation weight of client $i$, $|\xi^{(i)}| / \sum_{j\in[N]}|\xi^{(j)}|$ \\
$\omega^2$         & Upper bound on $\sum_{i \in \mathbb{G}_\tau}(w^{(i)})^2$ \\ \hline

\end{tabular}
\end{table}

\section{Convergence analyses}
\subsection{Lemmas}\label{appdx:proof-lemma}
\begin{lemma}(Bounded expectation of intra-group local updates)\label{lemma:Delta} Under Assumptions 1-3, for any group $\mathbb{G}_\tau$, we have 
    \begin{align}\label{ineq:lemma-1}
        \mathbb{E}[\|\bar{\Delta}_{\mathbb{G}_\tau}\|^2 ]\leq \Gamma^2
        + 2B^2\gamma^2 \mathbb{E}\left[ \left\|\nabla f(\mathbf{x}_{\pi(\tau)}) \right\|^2 \right] ,
    \end{align}
where $\Gamma^2:=B\gamma^2\sigma^2\omega^2
        + 4 L^2B^2\gamma^2
        + 4 B^2\gamma^2 (\tilde{\alpha}+\tilde{\beta})^2$.
\end{lemma}
\begin{proof}
By definition in Alg.~\ref{alg:PipeCycle}, we decompose the LHS of (\ref{ineq:lemma-1}) to have
\allowdisplaybreaks{
\begin{align*}
    &\mathbb{E}\big[\|\bar{\Delta}_{\mathbb{G}_\tau}\|^2\big] \nonumber \\
    &\overset{(a)}{=} \mathbb{E}\Bigg[\Bigg\| \sum_{i\in\mathbb{G}_\tau} \frac{|\xi_i|} {\sum_{j\in\mathbb{G}_\tau}|\xi_j|} \Delta_\tau^{(i)}\Bigg\|^2 \Bigg] \nonumber \\
    &= \gamma^2 \mathbb{E}\Bigg[ \Bigg\| \sum_{i\in\mathbb{G}_\tau} w^{(i)} \sum_{b=0}^{B-1} g_i (\mathbf{y}_{\tau,b}^{(i)})\Bigg\|^2 \Bigg] \nonumber \\
    &\overset{(b)}{=} \gamma^2 \mathbb{E}\Bigg[\Bigg\| \sum_{i\in\mathbb{G}_\tau} w^{(i)} \sum_{b=0}^{B-1} \left(g_i(\mathbf{y}_{\tau,b}^{(i)}) -\nabla f_i(\mathbf{y}_{\tau,b}^{(i)})\right)\Bigg\|^2 \Bigg] \nonumber \\
        &\quad+ \gamma^2 \mathbb{E}\Bigg[\Bigg\| \sum_{i\in\mathbb{G}_\tau} w^{(i)} \sum_{b=0}^{B-1} \nabla f_i (\mathbf{y}_{\tau,b}^{(i)})\Bigg\|^2 \Bigg]\nonumber \\
    &\overset{(c)}{\leq} B\gamma^2\sigma^2 \mathbb{E}\Bigg[\sum_{i\in\mathbb{G}_\tau} (w^{(i)})^2 \Bigg] \nonumber \\
        &\quad+ \gamma^2 \mathbb{E}\Bigg[ \Bigg\| \sum_{i\in\mathbb{G}_\tau} w^{(i)} \sum_{b=0}^{B-1} \Big(\nabla f(\mathbf{x}_{\pi(\tau)}) +\nabla f_i (\mathbf{y}_{\tau,b}^{(i)})\nonumber \\
            &\qquad-\nabla f(\mathbf{x}_{\pi(\tau)})\Big) \Bigg\|^2 \Bigg] \nonumber \\
    &\overset{(d)}{\leq} B\gamma^2\sigma^2\omega^2
        + 2\gamma^2 \mathbb{E}\left[ \left\| \sum_{i\in\mathbb{G}_\tau} w^{(i)} \sum_{b=0}^{B-1} \nabla f(\mathbf{x}_{\pi(\tau)}) \right\|^2 \right] \nonumber \\
        &\quad+ 2\gamma^2 \mathbb{E}\left[ \left\| \sum_{i\in\mathbb{G}_\tau} w^{(i)} \sum_{b=0}^{B-1} \left(\nabla f_i (\mathbf{y}_{\tau,b}^{(i)}) -\nabla f(\mathbf{x}_{\pi(\tau)})\right) \right\|^2 \right] \nonumber \\
    &\leq B\gamma^2\sigma^2\omega^2
        + 2B^2\gamma^2 \mathbb{E}\left[ \left\|\nabla f(\mathbf{x}_{\pi(\tau)}) \right\|^2 \right] \nonumber \\
        &\quad+ 2\gamma^2 \mathbb{E}\Bigg[ \Bigg\| \sum_{b=0}^{B-1} \Big(\nabla f_i (\mathbf{y}_{\tau,b}^{(i)}) -\nabla f_i(\mathbf{x}_{\pi(\tau)})  \nonumber \\
        &\qquad+\nabla f_i (\mathbf{x}_{\pi(\tau)}) -\nabla f(\mathbf{x}_{\pi(\tau)})\Big) \Bigg\|^2 \Bigg] \nonumber \\
    &\overset{(e)}{\leq} B\gamma^2\sigma^2\omega^2
        + 2B^2\gamma^2 \mathbb{E}\left[ \left\|\nabla f(\mathbf{x}_{\pi(\tau)}) \right\|^2 \right] \nonumber \\
        &\quad+ 2\gamma^2 \mathbb{E}\left[ \left\| \sum_{b=0}^{B-1} \left(\nabla f_i (\mathbf{y}_{\tau,b}^{(i)}) -\nabla f_i(\mathbf{x}_{\pi(\tau)}) \right)\right\|^2 \right] \nonumber \\
        &\quad +2\gamma^2 \mathbb{E}\left[ \left\|\sum_{b=0}^{B-1}\left(\nabla f_i (\mathbf{x}_{\pi(\tau)}) -\nabla f(\mathbf{x}_{\pi(\tau)})\right) \right\|^2 \right] \nonumber \\
    &\overset{(f)}{\leq} \underbrace{B\gamma^2\sigma^2\omega^2
        + 4 L^2B^2\gamma^2
        + 4 B^2\gamma^2 (\tilde{\alpha}+\tilde{\beta})^2}_{:=\Gamma^2} \nonumber \\
        &\quad+ 2B^2\gamma^2 \mathbb{E}\left[ \left\|\nabla f(\mathbf{x}_{\pi(\tau)}) \right\|^2 \right],
\end{align*}
where (a) applies (\ref{def:intra_group_Delta}); (b) results from the definition of variance; (c) uses Jensen's inequality and Assumption~\ref{assmp:sg} on the first term; (d) takes the definition of $\omega^2$ from (\ref{def:omega}) and separates the second term into two by applying $\|\mathbf{x}+\mathbf{y}\|^2\leq 2\|\mathbf{x}\|^2 +2\|\mathbf{y}\|^2$ for any $\mathbf{x}, \mathbf{y}$; (e) uses Young's inequality on the last term; (f) is due to rephrased Peter-Paul Inequality, following by Assumption~\ref{assmp:lipschitz} on the third term and Assumption~\ref{assmp:group-heterogeneity} and triangle inequality on the fourth term. For simplicity, the sum of all the terms independent on $\tau$ is denoted $\Gamma^2$.
} 
\end{proof}

\begin{lemma}(Bounded expectation of perturbation)\label{lemma:perturbation} 
For any $\tau$ and with Lemma~\ref{lemma:Delta}, the expectation of the perturbation is bounded by
\begin{align*}
    \mathbb{E}[\|\mathbf{e}_\tau\|^2] &\leq
    d_{\max}(G-1) \Gamma^2\nonumber \\
    &\hspace{0.5em}+2B^2\gamma^2 d_{\max}(G-1) \sum_{m\in\mathbb{A}'(\tau)} \mathbb{E}\big[ \|\nabla f(\mathbf{x}_{\pi(m)}) \|^2 \big] ,
\end{align*}
where $\mathbb{A}'(\tau) = \{1,\cdots,T\}\setminus\mathbb{A}(\tau)$ denotes a complement set composed of event epochs up to $\tau$ that are not in $\tau$'s ancestry chain.
\end{lemma}
\begin{proof}
The perturbation vector can be rewritten as
\[
    \mathbf{e}_\tau = \sum_{m\in\mathbb{A}'(\tau)} \bar{\Delta}_{\mathbb{G}_m} .
\]
We adopt the concept \emph{depth} that quantifies how many ancestor groups a group has gone through. An ancestry chain of group $\mathbb{G}_\tau$ can be written as $\{\tau, \pi(\tau), \pi^2(\tau), \cdots, \pi^{d_\tau}(\tau)\}$, where $d_\tau$ is the depth of this chain and depends on $\tau$. Since the proposed framework is operated over a finite horizon with available time slots ranging between $s\in(0, S_{\mathrm{tot}}]$, any $d_\tau$ is bounded to $d_{\max} = \lfloor \frac{S_{\mathrm{tot}}}{B} \rfloor$.
Therefore, under finite operation of the framework, we have $|\mathbb{A}'(\tau)|<d_{\max}(G-1)$ for any $\tau$ because the ancestry chain has at most $d_{\max}$ levels and each level can have at most $G-1$ contemporary events. Thus, we have
\begin{align}\label{ineq:perturbation_1}
    \|\mathbf{e}_\tau\|^2 &= \Bigg\| \sum_{m\in\mathbb{A}'(\tau)} \bar{\Delta}_{\mathbb{G}_m} \Bigg\|^2 \nonumber \\
    &\leq d_{\max}(G-1) \sum_{m\in\mathbb{A}'(\tau)} \big\|  \bar{\Delta}_{\mathbb{G}_m} \big\|^2 .
\end{align}
Taking expectation on both side of (\ref{ineq:perturbation_1}) and applying Lemma~\ref{lemma:Delta}, we have
\begin{align*}
    \mathbb{E}[\|\mathbf{e}_\tau\|^2] &\leq
    d_{\max}(G-1) \Gamma^2\nonumber \\
    &\hspace{0.5em}+2B^2\gamma^2 d_{\max}(G-1) \sum_{m\in\mathbb{A}'(\tau)} \mathbb{E}\big[ \|\nabla f(\mathbf{x}_{\pi(m)}) \|^2 \big] .
\end{align*}
\end{proof}

\subsection{Proof of Theorem~\ref{thm:0}}\label{appdx:proof-thm}
\begin{proof}
By definition of $\mathbf{z}$, the distance between two virtual global models observed at any event epoch $m$ and $n$ satisfying $0\leq m<n\leq T$ are
\allowdisplaybreaks{
\begin{align}\label{ineq:zmzn}
    \left\|\mathbf{z}_m - \mathbf{z}_n\right\|^2 = \Bigg\|\sum_{k=m+1}^{n} \bar{\Delta}_{\mathbb{G}_k}\Bigg\|^2 .
\end{align}}
Due to $L$-smoothness, we have
\allowdisplaybreaks{
\begin{align}\label{ineq:ztau+1_and_ztau}
    &f(\mathbf{z}_{\tau+1}) \nonumber \\
    &\leq f(\mathbf{z}_{\tau}) + \big\langle \nabla f(\mathbf{z}_{\tau}), \mathbf{z}_{\tau+1}-\mathbf{z}_{\tau} \big\rangle + \frac{L}{2}\|\mathbf{z}_{\tau+1}-\mathbf{z}_{\tau}\|^2 \nonumber \\
    &\overset{(a)}{=} f(\mathbf{z}_{\tau})
        + \big\langle \nabla f(\mathbf{z}_{\tau}), \bar{\Delta}_{\mathbb{G}_{\tau+1}} \big\rangle
        + \frac{L}{2}\|\bar{\Delta}_{\mathbb{G}_{\tau+1}}\|^2 \nonumber \\
    &\overset{(b)}{=} f(\mathbf{z}_{\tau})
        + \left\langle \nabla f(\mathbf{x}_{\pi(\tau+1)}), \bar{\Delta}_{\mathbb{G}_{\tau+1}} \right\rangle \nonumber \\
        &\quad+\left\langle \nabla f(\mathbf{z}_\tau)- \nabla f(\mathbf{x}_{\pi(\tau+1)}), \bar{\Delta}_{\mathbb{G}_{\tau+1}}\right\rangle
        + \frac{L}{2}\|\bar{\Delta}_{\mathbb{G}_{\tau+1}}\|^2 \nonumber \\
    &\overset{(c)}{\leq} f(\mathbf{z}_{\tau})
        + \left\langle \nabla f(\mathbf{x}_{\pi(\tau+1)}), \bar{\Delta}_{\mathbb{G}_{\tau+1}} \right\rangle\nonumber \\
        &\quad+ \left\| \nabla f(\mathbf{z}_\tau) - \nabla f(\mathbf{x}_{\pi(\tau+1)})\right\| \cdot \left\|\bar{\Delta}_{\mathbb{G}_{\tau+1}}\right\|
        + \frac{L}{2}\|\bar{\Delta}_{\mathbb{G}_{\tau+1}}\|^2 \nonumber \\
    &\overset{(d)}{\leq} f(\mathbf{z}_{\tau})
        + \left\langle \nabla f(\mathbf{x}_{\pi(\tau+1)}), \bar{\Delta}_{\mathbb{G}_{\tau+1}} \right\rangle \nonumber \\
        &\quad+ L\left\| \mathbf{z}_\tau - \mathbf{x}_{\pi(\tau+1)}\right\| \cdot \left\|\bar{\Delta}_{\mathbb{G}_{\tau+1}}\right\|
        + \frac{L}{2}\|\bar{\Delta}_{\mathbb{G}_{\tau+1}}\|^2 \nonumber \\
    &\overset{(e)}{\leq} f(\mathbf{z}_{\tau})
        + \left\langle \nabla f(\mathbf{x}_{\pi(\tau+1)}), \bar{\Delta}_{\mathbb{G}_{\tau+1}} \right\rangle \nonumber \\
        &\quad+ \frac{L}{2}\left\| \mathbf{z}_\tau - \mathbf{x}_{\pi(\tau+1)}\right\|^2
        + L \left\|\bar{\Delta}_{\mathbb{G}_{\tau+1}}\right\|^2
\end{align}
where (a) takes (\ref{expansion_ztau}); (b) uses the decomposition of $\bar{\Delta}_{\mathbb{G}_{\tau+1}}$, which is computed from the reference model $\mathbf{x}_{\pi(\tau+1)}$ in real implementation (i.e., $\bar{\Delta}_{\mathbb{G}_{\tau+1}} = \mathbf{x}_{\tau+1} - \mathbf{x}_{\pi(\tau+1)}$.); (c) comes from Cauchy-Schwarz inequality on the third term; (d) applies $L$-smoothness on the third term; (e) results from Young's inequality on the third term.
} 

The second term of the last RHS of (\ref{ineq:ztau+1_and_ztau}) involves a reference model update within single pipeline (group), of which the bound is shown as
\allowdisplaybreaks{
\begin{align}\label{ineq:inner_product}
    &\mathbb{E}\left[ \left\langle \nabla f(\mathbf{x}_{\pi(\tau+1)}), \bar{\Delta}_{\mathbb{G}_{\tau+1}} \right\rangle \right] \nonumber \\
    &=\mathbb{E}\left[ \left\langle \nabla f(\mathbf{x}_{\pi(\tau+1)}),
        \mathbb{E}\left[\bar{\Delta}_{\mathbb{G}_{\tau+1}} | \mathbf{x}_{\pi(\tau+1)}\right] \right\rangle \right] \nonumber \\
    &\overset{(a)}{\approx} \mathbb{E}\Big[ \Big\langle \nabla f(\mathbf{x}_{\pi(\tau+1)}),
        -B\gamma \sum_{i\in\mathbb{G}_{\tau+1}}\nabla f_i(\mathbf{x}_{\pi(\tau+1)}) \Big\rangle \Big] \nonumber \\
    &\overset{(b)}{\leq} -\frac{B\gamma}{2}\mathbb{E}[\|\nabla f(\mathbf{x}_{\pi(\tau+1)})\|^2] \nonumber \\
        &\quad+ \frac{B\gamma}{2} \Big\| f(\mathbf{x}_{\pi(\tau+1)}) -\sum_{i\in\mathbb{G}_{\tau+1}}\nabla f_i(\mathbf{x}_{\pi(\tau+1)}) \Big\|^2 \nonumber \\
    &\overset{(c)}{\leq} -\frac{B\gamma}{2}\mathbb{E}[\|\nabla f(\mathbf{x}_{\pi(\tau+1)})\|^2] +\frac{\tilde{\beta}^2B\gamma}{2},
\end{align}
where (a) uses Assumption \ref{assmp:sg}; (b) takes the rephrased definition of inner product $-\langle \mathbf{x},\mathbf{y}\rangle=\frac{1}{2}\left(-\|\mathbf{x}\|^2-\|\mathbf{y}\|^2 +\|\mathbf{x}-\mathbf{y}\|^2\right)$; and (c) adopts Assumption \ref{assmp:group-heterogeneity}.
} 

The third term of the last RHS of (\ref{ineq:ztau+1_and_ztau}) without coefficient can be bounded as follows.
\allowdisplaybreaks{
\begin{align}\label{ineq:ztau_and_xtau+1}
    &\left\| \mathbf{z}_\tau - \mathbf{x}_{\pi(\tau+1)}\right\|^2 \nonumber \\
    &=\left\| \mathbf{z}_\tau - \mathbf{z}_{\pi(\tau+1)}
        +\mathbf{z}_{\pi(\tau+1)} -\mathbf{x}_{\pi(\tau+1)}\right\|^2 \nonumber \\
    &=\left\| \mathbf{z}_\tau - \mathbf{z}_{\pi(\tau+1)}
        +\mathbf{e}_{\pi(\tau+1)}\right\|^2 \nonumber \\
    &\overset{(a)}{\leq} 2\left\| \mathbf{z}_\tau - \mathbf{z}_{\pi(\tau+1)}\right\|^2
        +2\left\|\mathbf{e}_{\pi(\tau+1)}\right\|^2 \nonumber \\
    &\overset{(b)}{\leq} 2\Bigg\| \sum_{m=\pi(\tau+1)+1}^\tau \bar{\Delta}_{\mathbb{G}_m}\Bigg\|^2
        +2\left\|\mathbf{e}_{\pi(\tau+1)}\right\|^2 \nonumber \\
    &\overset{(c)}{\leq} 2(G-1)\sum_{m=\pi(\tau+1)+1}^\tau\left\|  \bar{\Delta}_{\mathbb{G}_m}\right\|^2
        +2\left\|\mathbf{e}_{\pi(\tau+1)}\right\|^2,
\end{align}
where (a) is from Young's inequality; (b) uses (\ref{ineq:zmzn}); (c) takes Jensen's Inequality. Note that there can exist at most $G-1$ instances in the event epoch range $[\pi(\tau+1)+1,\ \tau]$.
}

Applying (\ref{ineq:inner_product}) and (\ref{ineq:ztau_and_xtau+1}) to (\ref{ineq:ztau+1_and_ztau}) and taking expectations, we have
\allowdisplaybreaks{
\begin{align}\label{ineq_2:ztau+1_and_ztau}
    &\mathbb{E}[f(\mathbf{z}_{\tau+1})] \nonumber\\
    &\leq \mathbb{E}[f(\mathbf{z}_\tau)] 
        -\frac{B\gamma}{2}\mathbb{E}[\|\nabla f(\mathbf{x}_{\pi(\tau+1)})\|^2]
        +\frac{\tilde{\beta}^2B\gamma}{2} \nonumber\\
        &\quad+ L(G-1)\mathbb{E}\Bigg[ \sum_{m=\pi(\tau+1)+1}^\tau\left\|  \bar{\Delta}_{\mathbb{G}_m}\right\|^2\Bigg] \nonumber\\
        &\quad+L\mathbb{E}\left [\left\|\mathbf{e}_{\pi(\tau+1)}\right\|^2 \right] 
        +L \mathbb{E}\left [\left\|\bar{\Delta}_{\mathbb{G}_{\tau+1}}\right\|^2 \right] \nonumber \\
    &\overset{(a)}{\leq} \mathbb{E}[f(\mathbf{z}_\tau)] 
        -\frac{B\gamma}{2}\mathbb{E}[\|\nabla f(\mathbf{x}_{\pi(\tau+1)})\|^2]
        +\frac{\tilde{\beta}^2B\gamma}{2} \nonumber \\
        &\quad + L(G-1) \Big( \Gamma^2
        + 2B^2\gamma^2 \sum_{m=\pi(\tau+1)+1}^\tau \mathbb{E}\big[ \|\nabla f(\mathbf{x}_{\pi(m)}) \|^2 \big] \Big) \nonumber\\
        &\quad+L\mathbb{E}\left [\left\|\mathbf{e}_{\pi(\tau+1)}\right\|^2 \right]\nonumber \\
        &\quad+L\left( \Gamma^2
        + 2B^2\gamma^2 \mathbb{E}\left[ \left\|\nabla f(\mathbf{x}_{\pi(\tau+1)}) \right\|^2 \right] \right)
        \nonumber \\
    &\overset{(b)}{\leq} \mathbb{E}[f(\mathbf{z}_\tau)] 
        -\frac{B\gamma}{2}\mathbb{E}[\|\nabla f(\mathbf{x}_{\pi(\tau+1)})\|^2]
        +\frac{\tilde{\beta}^2B\gamma}{2} \nonumber \\
        &\quad + L(G-1)\Gamma^2 \nonumber \\
        &\quad + 2B^2\gamma^2L(G-1)  \sum_{m=\pi(\tau+1)+1}^\tau \mathbb{E}\left[ \left\|\nabla f(\mathbf{x}_{\pi(m)}) \right\|^2 \right]  \nonumber \\ 
        &\quad+ d_{\max}L(G-1) \Gamma^2 \nonumber\\
        &\quad+ 2B^2\gamma^2 d_{\max}L(G-1) \sum_{n\in\mathbb{A}'(\pi(\tau+1))} \mathbb{E}\left[ \left\|\nabla f(\mathbf{x}_{\pi(n)}) \right\|^2 \right] \nonumber \\
        &\quad+ L \Gamma^2
        + 2B^2\gamma^2L \mathbb{E}\left[ \left\|\nabla f(\mathbf{x}_{\pi(\tau+1)}) \right\|^2 \right]
        \nonumber \\
    &= \mathbb{E}[f(\mathbf{z}_\tau)] 
        -\left( \frac{B\gamma}{2}-2B^2\gamma^2L\right) \mathbb{E}\left[ \left\|\nabla f(\mathbf{x}_{\pi(\tau+1)}) \right\|^2 \right]
         \nonumber \\ 
        &\quad   
        + 2B^2\gamma^2L(G-1)  \sum_{m=\pi(\tau+1)+1}^\tau \mathbb{E}\left[ \left\|\nabla f(\mathbf{x}_{\pi(m)}) \right\|^2 \right]  \nonumber \\  
        &\quad
        + 2B^2\gamma^2 d_{\max}L(G-1) \sum_{n\in\mathbb{A}'(\pi(\tau+1))} \mathbb{E}\left[ \left\|\nabla f(\mathbf{x}_{\pi(n)}) \right\|^2 \right]  \nonumber\\ 
        &\quad+ \underbrace{ L\big((d_{\max}+1)(G-1)+1\big) \Gamma^2
        +\frac{\tilde{\beta}^2B\gamma}{2}}_{:=\Lambda^2} 
\end{align}
where (a) is from applying Lemma \ref{lemma:Delta} to the terms with $\mathbb{E}[\|\bar{\Delta}_{\mathbb{G}_\cdot}\|^2]$; (b) results from applying Lemma~\ref{lemma:perturbation}.
}
Summing all sequence $\mathbb{E}[f(\mathbf{z}_{\tau+1})]-\mathbb{E}[f(\mathbf{z}_\tau)]$ of (\ref{ineq_2:ztau+1_and_ztau}) from $\tau=0$ to $T-1$, we have
\allowdisplaybreaks{
\begin{align}\label{ineq_3:ztau+1_and_ztau}
    &\mathbb{E}[f(\mathbf{z}_T)]-f(\mathbf{z}_0)  \nonumber \\
        &\leq -\left( \frac{B\gamma}{2} -2B^2\gamma^2L \right) \sum_{m=1}^T\mathbb{E}\left[ \left\|\nabla f(\mathbf{x}_{\pi(m)}) \right\|^2 \right] \nonumber \\ 
        &\quad\overset{(a)}{+} 2B^2\gamma^2L(G-1)^2  \sum_{n=1}^T \mathbb{E}\left[ \left\|\nabla f(\mathbf{x}_{\pi(n)}) \right\|^2 \right]  \nonumber \\  
        &\quad\overset{(b)}{+} 2B^2\gamma^2 d_{\max}^2L(G-1)^2 \sum_{k=1}^T \mathbb{E}\left[ \left\|\nabla f(\mathbf{x}_{\pi(k)}) \right\|^2 \right]
        +\Lambda^2 T \nonumber\\ 
    &= -\underbrace{\left( \frac{B\gamma}{2} -2B^2\gamma^2L -2B^2\gamma^2 (d_{\max}^2+1)L(G-1)^2 \right)}_{:=\zeta} \nonumber\\
    &\qquad \cdot \sum_{\tau=1}^T\mathbb{E}\big[ \|\nabla f(\mathbf{x}_{\pi(\tau)}) \|^2 \big]  +\Lambda^2 T .
\end{align}
The bound in (a) and (b) is derived from counting how many times each $\mathbb{E}[\|\nabla f(\mathbf{x}_{\pi(\cdot)})\|^2]$ appears on the RHS. From the $\bar{\Delta}$'s sum (a), event $n$ appears when some group $\mathbb{G}_{\tau+1}$ was actively local training at the time $n$ completed. Since at most $G-1$ groups can coexist at any moment, each $n$ appears maximum $G-1$ times. From the perturbation sum (b), event $k$ appears in $\mathbb{A}'\left(\pi(\tau+1)\right)$ when event $k$ lies on a chain that is different from the one containing event $\pi(\tau+1)$. Since the relay structure forms at most $G$ parallel chains with at most $d_{\max}$ of length, event $k$ can appear in $\mathbb{A}'\left(\pi(\tau+1)\right)$ for at most $d_{\max}(G-1)$ across all events $\tau\in[0,T)$; there are at most $d_{\max}$ ancestry levels, where each level can have at most $G-1$ other chains.
Note that $\zeta$ is always positive given any learning rate that satisfies $\gamma<\frac{1}{4B L\left(1+ (d_{\max}^2+1)(G-1) \right)}$.
}

Finally, by rephrasing (\ref{ineq_3:ztau+1_and_ztau}) and dividing both sides into $T$, we obtain
\allowdisplaybreaks{
\begin{align}
    \min_{\tau} \mathbb{E}\left[ \left\|\nabla f(\mathbf{x}_{\pi(\tau)}) \right\|^2 \right] 
    &\leq \frac{1}{T}\sum_{\tau=1}^T\mathbb{E}\left[ \left\|\nabla f(\mathbf{x}_{\pi(\tau)}) \right\|^2 \right] \nonumber\\
    &\leq \frac{f(\mathbf{z}_0) - f^*}{\zeta T} + \frac{\Lambda^2}{\zeta}.
\end{align}
}
\end{proof}

\subsection{Proof of Corollary~\ref{corol:0}}\label{appdx:proof-corol}
The total number of aggregation events $T$ cannot be fixed before running the algorithm because it depends on how fast groups are formed and completed. In other words, $T$ depends on the pipeline structure ($G, B$), the scale of distributed network ($N$), and energy availability ($\delta$).

After running the framework for substantial slots, the system reaches some status like steady-state, which is a long-run average per each client in the strict sense.
Once a user has just finished local training at slot $s$, its expected residual battery level is
\[
    \mathbb{E}[E_{s_B}^{(i)}] = \mathbb{E}[E_{s_0}^{(i)}] - (B+1) + B\delta ,
\]
where $s_0=s-B$ is an index of time slot at which user $i$ started local training. The calculation is based on the algorithm: $E_{s_0}^{(i)}\geq B+1$ is the initial battery status criterion, the client spent $B+1$ units ($B$ for training, $1$ for transmission), and gained $\Bin(B,\delta)$ units from charging during $B$ slots.

The expected number of slots to be ready for the next participation is
\begin{align*}
    \mathbb{E}[s_\text{recharge}]
    &= \frac{B+1 -\mathbb{E}[E_{s_B}^{(i)}]}{\delta} \\
    &= \frac{B+1 -\mathbb{E}[E_{s_0}^{(i)}] +B+1 - B\delta}{\delta} .
\end{align*}
In the worst (most conservative) case where $\mathbb{E}[E_{s_0}^{(i)}]=B+1$, we have the minimum expectation of the number of slots between two successive user participations:
\[
    \mathbb{E}[\underline{s}_\text{recharge}] = \frac{B+1 - B\delta}{\delta} ,
\]
which is equal to a client cycling period. This is equal to the most conservatively expected time for each user to wait from group participation to the next participation. 

From client cycling period, we can define event rate as the average number of aggregation events per slot when the system achieves quasi-steady state.
In the long-run average, each of $N$ clients participates once every $\underline{s}_\text{recharge}$ idle slots. Considering that a group has at most $n_G$ members, the energy-determined event rate is $\frac{N\delta}{n_G(B+1-B\delta)}$. Meanwhile, a system can achieve maximum possible rate at $\frac{G}{B}$, when all users always have abundant batteries to join a group.
Thus, the effective event rate is the binding bottleneck:
\begin{align}\label{eq:event_rate}
    r = \min\left(\frac{N\delta}{n_G(B+1-B\delta)}, \frac{G}{B} \right) .
\end{align}

The total number of aggregation events $T\approx S_{\mathrm{tot}}\cdot r$ for sufficiently large $S_{\mathrm{tot}}$.

\subsection{Proof of Remark on page~\pageref{optimal_G}}\label{appdx:proof-remark}
\begin{proof}
Having $n_G=\lfloor \frac{N}{2.5G} \rfloor$ in (\ref{eq:event_rate}), the event rate $r$ can be redefined as follows.
\begin{equation}
    r=
    \begin{cases}
        \frac{5G\delta}{2(B+1-B\delta)}, &\text{if }\delta\leq\frac{2(B+1)}{7B} \\
        \frac{G}{B}, &\text{otherwise.}
    \end{cases}
\end{equation}
Note that the event rate is proportional to $G$ regardless of the value of $\delta$. The stationarity gap in Theorem~\ref{thm:0} is thus bounded by $\mathcal{O}(\frac{f(\mathbf{x}_0) - f^*}{h(G)})$, where $h(G)=G\zeta(G)=\frac{B\gamma}{2}G -2B^2\gamma^2LG -2B^2\gamma^2 (d_{\max}^2+1)LG(G-1)^2$. We want to minimize the entire RHS over $G$ to find the tightest possible bound. Given that the asymptotic floor becomes negligible for substantially large $S_{\mathrm{tot}}$, finding optimal $G$ is equivalent to finding $G$ that maximizes $h(G)$. Since $\frac{dh(G)}{dG}$ is a downward parabola with two roots that satisfy $\frac{dh(G)}{dG}=0$, the larger root is the local maximum.    
\end{proof}


\section{Detailed Experiment Settings}\label{appdx:exp-settings}
To ensure fair comparison across algorithms and to eliminate variance attributable to data partitioning, we run 10 trials per configuration with pre-generated and pre-saved data splits. Within each trial, the training partition across clients and the held-out test set are fixed and reused across all algorithm and group-size combinations; the partition changes only when the trial index changes. Different trials use different random seeds for both data partitioning and run-time stochasticity (mini-batch sampling and hub selection). This protocol prevents an algorithm from appearing artificially favorable or unfavorable due to a lucky or unlucky random draw, and isolates the performance differences to the algorithms themselves. We report the mean across the 10 trials.

All experiments were implemented in PyTorch 2.1.2 with CUDA 12.1 and run on NVIDIA Tesla T4 GPUs (15\,GB memory) on the Alvis cluster, with Intel Xeon Gold 6226R CPUs at 2.90\,GHz.

\newpage
\section{Pseudo algorithm}\label{appdx:pseudo-alg}

Alg.~\ref{alg:PipeCycle} demonstrates the pseudo-algorithm of PipeCycle.



\begin{algorithm}[!htb]
\caption{PipeCycle: Pipelined Energy-Harvesting FedAvg with Battery-aware Group Formation.}
\label{alg:PipeCycle}
\DontPrintSemicolon
\KwIn{$S_{\mathrm{tot}}$: slots; $[N]$: users; $n_G$: max group size; $B$: local duration; $R$: formation interval; $\delta$: charge prob.; $E_{\max}$: max battery; $\gamma$: learning rate.}
\KwOut{Server model $\mathbf{x}$}

$\mathbf{x} \leftarrow \mathbf{x}_0$; $v \leftarrow 0$; $\mathcal{Q} \leftarrow \emptyset$; $\mathcal{H} \leftarrow \emptyset$; $s_{\mathrm{last}} \leftarrow -R$\;

\For{$s = 0$ \KwTo $S_{\mathrm{tot}}-1$}{

    \tcp{Battery charging}
    \ForEach{$i \in [N]$ with prob.\ $\delta$}
        {$E_s^{(i)} \leftarrow \min(E_s^{(i)}+1,\, E_{\max})$}

    \tcp{Group formation (longest-idle-first)}
    \If{$s - s_{\mathrm{last}} \ge R$}{
        $\mathcal{A} \leftarrow \{i \in [N] : E_s^{(i)} \ge B{+}1,\; \text{idle},\; \nexists\,\Delta^{(i)}\}$\;
        $\mathbb{G} \leftarrow \text{top-}n_G$ from $\mathcal{A}$ by longest idle time\;
        \eIf{$\mathcal{H} \neq \emptyset$\tcp*[f]{relay from prev.\ hub}}{
            $(\hat{\imath},\hat{\mathbf{x}}) \leftarrow \mathcal{H}.\mathrm{dequeue}()$;\;
            $\mathbf{x}_{\mathrm{ref}} \leftarrow \hat{\mathbf{x}}$;\;
            $\mathbf{x} \leftarrow \hat{\mathbf{x}}$;\; $v \mathrel{+}= 1$;\; $E_s^{(\hat{\imath})} \mathrel{-}= 1$\;
        }{
            $\mathbf{x}_{\mathrm{ref}} \leftarrow \mathbf{x}$\tcp*[r]{seed from server}
        }
        \lForEach{$j \in \mathbb{G}$}{$\mathbf{x}^{(j)} \leftarrow \mathbf{x}_{\mathrm{ref}}$;}
        $\mathcal{Q}.\mathrm{enqueue}(\mathbb{G})$;\; $s_{\mathrm{last}} \leftarrow s$\;        
    }

    \tcp{Flush stagnated pipeline}
    \If{$\mathcal{Q} = \emptyset$ \textbf{and} $\mathcal{H} \neq \emptyset$}{
        $(\hat{\imath},\hat{\mathbf{x}}) \leftarrow \mathcal{H}.\mathrm{dequeue}()$;\;
        $\mathbf{x} \leftarrow \hat{\mathbf{x}}$;\; $v \mathrel{+}= 1$;\; $E^{(\hat{\imath})} \mathrel{-}= 1$\;
    }

    \tcp{Local training (parallel across groups)}
    \ForEach{$i$ in any $\mathbb{G} \in \mathcal{Q}$ with remaining slots $> 0$}{
        $\mathbf{x}^{(i)} \leftarrow \mathbf{x}^{(i)} - \gamma \nabla f_i(\mathbf{x}^{(i)};\xi^{(i)})$;\; $E^{(i)} \mathrel{-}= 1$\;
    }

    \tcp{Intra-group aggregation}
    \ForEach{$\mathbb{G} \in \mathcal{Q}$ with all members finished}{
        $\bar{\Delta}_{\mathbb{G}} \leftarrow \textstyle\sum_{i \in \mathbb{G}} \tfrac{n_i}{\sum_j n_j}(\mathbf{x}^{(i)} - \bar{\mathbf{x}}^{(i)})$\tcp*[r]{weighted averaging}
        Pick hub $\hat{\imath}_{\mathbb{G}} \in \mathbb{G}$ uniformly at random.\;
        $\hat{\mathbf{x}}_{\mathbb{G}} \leftarrow \mathbf{x}^{(\hat{\imath}_{\mathbb{G}})} + \bar{\Delta}_{\mathbb{G}}$\;
        \ForEach{$i \in \mathbb{G} \setminus \{\hat{\imath}_{\mathbb{G}}\}$}
            {send $\Delta^{(i)}$ to hub;\;
            $E^{(i)} \mathrel{-}= 1$}
        $\mathcal{H}.\mathrm{enqueue}(\hat{\imath}_{\mathbb{G}},\hat{\mathbf{x}}_{\mathbb{G}})$;\; $\mathcal{Q}.\mathrm{remove}(\mathbb{G})$\;
    }
}
\Return{$\mathbf{x}$}
\end{algorithm}

\end{document}